%% file: arxiv_main.tex
\documentclass{article}

\usepackage{arxiv}

\usepackage[utf8]{inputenc} 
\usepackage[T1]{fontenc}    
\usepackage{hyperref}       
\usepackage{url}            
\usepackage{booktabs}       
\usepackage{amsfonts}       
\usepackage{nicefrac}       
\usepackage{microtype}      
\usepackage{lipsum}		
\usepackage{graphicx}
\usepackage{natbib}
\usepackage{doi}

\input{math_commands.tex}

\usepackage{hyperref}
\usepackage{url}
\usepackage{times}
\usepackage{algorithm}
\usepackage[normalem]{ulem}
\useunder{\uline}{\ul}{}
\usepackage{wrapfig}
\usepackage{booktabs}
\usepackage{multirow}
\usepackage{graphicx}
\usepackage{caption}
\usepackage{array}
\usepackage{longtable}
\newcolumntype{C}[1]{>{\centering\arraybackslash}p{#1}}
\usepackage{afterpage}
\usepackage{algorithm}
\usepackage[noend]{algpseudocode}

\usepackage{cancel}
\usepackage{subcaption}
\usepackage{color}
\usepackage[table,xcdraw]{xcolor}
\newcommand{\eat}[1]{}

\newcommand{\yu}[1]{{\color{red} [Yu: #1] }}
\newcommand{\icml}[1]{{#1}}
\newcommand{\sva}[1]{{[\color{olive} {\color{purple}\textbf{Shivvrat}}: #1] }}

\newcommand{\old}[1]{{\color{red} [Older Version (to be removed): #1]}}

\newcommand{\vibhav}[1]{{\color{orange} [Vibhav: #1] }}

\renewcommand{\sva}[1]{#1}
\renewcommand{\old}[1]{}

\newcommand{\cnn}[1]{CNN}
\newcommand{\mrfgs}[1]{DRF - GS}
\newcommand{\ilp}[1]{DRF - ILP}
\newcommand{\ijgp}[1]{DRF - IJGP}
\newcommand{\dnlr}[1]{DDN - LR - Pipeline}
\newcommand{\dnlrjoint}[1]{DDN - LR - Joint}
\newcommand{\dnnn}[1]{DDN - MLP - Pipeline}
\newcommand{\dnnnjoint}[1]{DDN - MLP - Joint}

\renewcommand{\cite}{\citep}

\title{Deep Dependency Networks and Advanced Inference Schemes for Multi-Label Classification}

\author{Shivvrat Arya \\
	Department of Computer Science\\
	The University of Texas at Dallas\\
	Richardson, TX 75252 \\
	\texttt{shivvrat.arya@utdallas.edu}
	\And
	Yu Xiang \\
	Department of Computer Science\\
	The University of Texas at Dallas\\
	Richardson, TX 75252 \\
	\texttt{yu.xiang@utdallas.edu} 
 	\And
	Vibhav Gogate \\
	Department of Computer Science\\
	The University of Texas at Dallas\\
	Richardson, TX 75252 \\
\texttt{vibhav.gogate@utdallas.edu}  
}

\date{}

\renewcommand{\shorttitle}{Deep Dependency Networks and Advanced Inference Schemes for MLC}

\hypersetup{
pdftitle={Deep Dependency Networks and Advanced Inference Schemes for Multi-Label Classification},
pdfsubject={cs.AI, cs.LG},
pdfauthor={Shivvrat Arya, Yu Xiang, Vibhav Gogate},
pdfkeywords={Multi-label Classification, Probabilistic Graphical Models, Multi-label Action Classification, Multi-label Image Classification, Dependency Networks, Probabilistic Modeling},
}
\begin{document}
\maketitle
\begin{abstract}
\input{content/0-abstract}

\end{abstract}

\section{INTRODUCTION}
\label{sec:intro}

\input{content/new-intro}

\section{PRELIMINARIES}
\label{sec:prelim}
\input{content/3-preliminary}

\section{TRAINING DEEP DEPENDENCY NETWORKS}
\label{sec:ddn}
\input{content/ddn}

\section{MPE INFERENCE IN DDNs}
\label{sec:mpe_inf}
\input{content/inference}


\section{EXPERIMENTAL EVALUATION}
\label{sec:experiments}
\input{content/6-experiment}

\section{RELATED WORK}
\label{section:related_work}
\input{content/2-related-work}

\section{CONCLUSION AND FUTURE WORK}
\label{sec:conclusion}
\input{content/7-conclusion}

\section*{ACKNOWLEDGMENTS}
This work was supported in part by the DARPA Perceptually-Enabled Task Guidance (PTG) Program under contract number HR00112220005, by the DARPA Assured Neuro Symbolic Learning and Reasoning (ANSR) under contract number HR001122S0039 and by the National Science Foundation grant IIS-1652835.
\bibliography{main}

\clearpage
\input{content/appendix}

\end{document}

%% file: math_commands.tex
\usepackage{amsmath,amsfonts,bm}

\def\eqref#1{equation~\ref{#1}}

\def\1{\bm{1}}

\def\rve{{\mathbf{e}}}

\def\rvv{{\mathbf{v}}}

\def\rvx{{\mathbf{x}}}

\def\rvz{{\mathbf{z}}}

\def\rmD{{\mathbf{D}}}
\def\rmE{{\mathbf{E}}}

\def\rmV{{\mathbf{V}}}

\def\rmX{{\mathbf{X}}}

\DeclareMathAlphabet{\mathsfit}{\encodingdefault}{\sfdefault}{m}{sl}
\SetMathAlphabet{\mathsfit}{bold}{\encodingdefault}{\sfdefault}{bx}{n}

\def\gG{{\mathcal{G}}}

\newcommand{\R}{\mathbb{R}}

\DeclareMathOperator*{\argmax}{arg\,max}

%% file: content/0-abstract.tex
We present a unified framework called deep dependency networks (DDNs) that combines dependency networks and deep learning architectures for multi-label classification, with a particular emphasis on image and video data. The primary advantage of dependency networks is their ease of training, in contrast to other probabilistic graphical models like Markov networks. In particular, when combined with deep learning architectures, they provide an intuitive, easy-to-use loss function for multi-label classification. A drawback of DDNs compared to Markov networks is their lack of advanced inference schemes, necessitating the use of Gibbs sampling. To address this challenge, we propose novel inference schemes based on local search and integer linear programming for computing the most likely assignment to the labels given observations. We evaluate our novel methods on three video datasets (Charades, TACoS, Wetlab) and three image datasets (MS-COCO, PASCAL VOC, NUS-WIDE), comparing their performance with (a) basic neural architectures and (b) neural architectures combined with Markov networks equipped with advanced inference and learning techniques. Our results demonstrate the superiority of our new DDN methods over the two competing approaches.

%% file: content/new-intro.tex
In this paper, we focus on the multi-label classification (MLC) task and more specifically, on its two notable instantiations, multi-label action classification (MLAC) for videos and multi-label image classification (MLIC). At a high level, given a pre-defined set of labels (or actions) and a test example (video or image), the goal is to assign each test example to a subset of labels.
It is well known that MLC is notoriously difficult because, in practice, the labels are often correlated, and thus, predicting them independently may lead to significant errors. Therefore, most advanced methods explicitly model the relationship or dependencies between the labels, using either probabilistic techniques \cite{wang2008generative,rossiProceedingsTwentyThirdInternational2013b, antonucci2013ensemble, wangEnhancingMultilabelClassification2014a, Tan_2015_CVPR,pmlr-v52-dimauro16} or non-probabilistic/neural methods  \cite{kongMultilabelClassificationMining2013a,papagiannopoulouDiscoveringExploitingDeterministic2015a,chenLearningSemanticSpecificGraph2019,chenMultiLabelImageRecognition2019,wang2021semi,nguyenModularGraphTransformer2021,wangNovelReasoningMechanism2021a,liuQuery2LabelSimpleTransformer2021, quMultilayeredSemanticRepresentation2021,liu_2022_text,zhou_2023_double, weng_2023_learning}.

\begin{figure}[t]
\vskip 0.2in
        \begin{center}
        \centerline{\includegraphics[width=0.99\columnwidth]{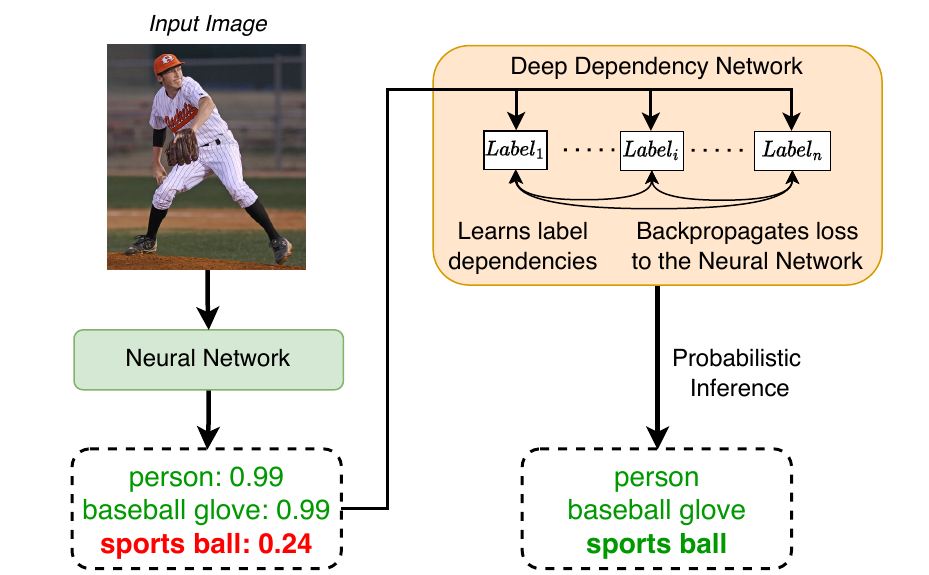}}
        \caption{Illustrating improvements from our new inference schemes for DDNs. The DDN learns relationships between labels, and the inference schemes reason over them to accurately identify concealed objects, such as  \textbf{sports ball}.}
        \label{fig:ddn_intro}
        \end{center}
\vspace{-8mm}
\end{figure}

To this end, motivated by approaches that combine probabilistic graphical models (PGMs) with neural networks (NNs)  \cite{krishnanDeepKalmanFilters2015, johnsonComposingGraphicalModels2017}, we jointly train a hybrid model, termed \textit{deep dependency networks} (DDNs) \cite{guo_2020_deep}, as illustrated in Figure \ref{fig:ddn_intro}. In a Deep Dependency Network (DDN), a conditional dependency network sits on top of a neural network. The underlying neural network transforms input data (e.g., an image) into a feature set. The dependency network \cite{heckerman2000dependency} then utilizes these features to establish a local conditional distribution for each label, considering not only the features but also the other labels. Thus, at a high level, a DDN is a \textit{neuro-symbolic model} where the neural network extracts features from data and the dependency network acts as a symbolic counterpart, learning the weighted constraints between the labels. 

However, a limitation of DDNs is that they rely on naive techniques such as Gibbs sampling and mean-field inference for probabilistic reasoning and lack advanced probabilistic inference techniques \cite{lowd_2011_mean,lowdClosedFormLearningMarkov2012a}. This paper addresses these limitations by introducing sophisticated inference schemes tailored for the Most Probable Explanation (MPE) task in DDNs, which involves finding the most likely assignment to the unobserved variables given observations. In essence, a solution to the MPE task, when applied to a probabilistic model defined over labels and observed variables, effectively solves the multi-label classification problem.

More specifically, we propose two new methods for MPE inference in DDNs. Our first method uses a random-walk based local search algorithm. Our second approach uses a piece-wise approximation of the log-sigmoid function to convert the non-linear MPE inference problem in DDNs into an integer linear programming problem. The latter can then be solved using off-the-shelf commercial solvers such as Gurobi \cite{gurobi}.

We evaluate DDNs equipped with our new MPE inference schemes and trained via joint learning on three video datasets: Charades \cite{charades_paper}, TACoS \cite{tacos:regnerietal:tacl}, and Wetlab \cite{wetlab:Naim2014UnsupervisedAO}, and three image datasets: MS-COCO \cite{linMicrosoftCOCOCommon2015}, PASCAL VOC 2007 \cite{everinghamPascalVisualObject2010}, and NUS-WIDE \cite{chuaNUSWIDERealworldWeb2009a}. We compare their performance to two categories of models: (a) basic neural networks (without dependency networks) and (b) hybrids of Markov random fields (MRFs), an undirected PGM, and neural networks equipped with sophisticated reasoning and learning algorithms.  Specifically, we employ three advanced approaches: (1) iterative join graph propagation (IJGP) \cite{mateescu2010join}, a type of generalized Belief propagation method \cite{yedidiaGeneralizedBeliefPropagation2000} for marginal inference, (2) integer linear programming (ILP) based techniques for computing most probable explanations (MPE) and (3) a well-known structure learning method based on logistic regression with $\ell_1$-regularization \cite{NIPS2006_lee_a4380923,NIPS2006_Wainwright_86b20716} for pairwise MRFs.

Via a detailed experimental evaluation, we found that, generally speaking, the MRF+NN hybrids outperform NNs, as measured by metrics such as Jaccard index and subset accuracy, especially when advanced inference methods such as IJGP are employed. Additionally, DDNs, when equipped with our novel MILP-based MPE inference approach, often outperform both MRF+NN hybrids and NNs. This enhanced performance of DDNs with advanced MPE solvers is likely attributed to their superior capture of dense label interdependencies, a challenge for MRFs. Notably, MRFs rely on sparsity for efficient inference and learning.

%% file: content/3-preliminary.tex
A \textbf{log-linear model} or a \textbf{Markov random field} (MRF), denoted by $\mathcal{M}$, is an undirected probabilistic graphical model \cite{koller2009probabilistic} that is widely used in many real-world domains for representing and reasoning about uncertainty. MRFs are defined as a triple $\langle \rmX, \mathcal{F}, \Theta \rangle$ where $\rmX{}=\{X_1,\ldots,X_n\}$ is a set of Boolean random variables, $\mathcal{F}=\{f_1,\ldots,f_m\}$ is a set of features such that each feature $f_i$ (we assume that a feature is a Boolean formula) is defined over a subset $\rmD_i$ of $\rmX$, and $\Theta=(\theta_1,\ldots,\theta_m)$ are real-valued weights or parameters, namely $\forall \theta_i \in \Theta;\; \theta_i \in \R$ such that each feature $f_i$ is associated with a parameter $\theta_i$. $\mathcal{M}$ represents the following probability distribution:
\begin{equation}
    P(\rvx)=\frac{1}{Z(\Theta)} \exp \left\{\sum_{i=1}^m \theta_i f_i\left(\rvx_{\rmD_i}\right)\right\}
\label{eq:log_linear_pd}
\end{equation}
where $\rvx$ is an assignment of values to all variables in $\rmX$, $\rvx_{\rmD_i}$ is the projection of $\rvx$ on the variables $\rmD_i$ of $f_i$, $f_i(\rvx_{\rmD_i})$ is an \textit{indicator function} that equals $1$ when $\rvx_{\rmD_i}$ evaluates $f_i$ to \texttt{True} and is $0$ otherwise, and $Z(\Theta)$ is the normalization constant called the \textit{partition function}.

\old{We focus on three tasks over MRFs: (1) structure learning, which involves determining the features and parameters from training data; (2) posterior marginal inference, which is the task of computing the marginal probability distribution over each variable in the network given evidence (evidence is defined as an assignment of values to a subset of variables); and (3) finding the most likely assignment to all the non-evidence variables given evidence (this task is often called maximum-a-posteriori or MAP inference in short) \sva{Should we change this to MPE? We say later that we solve MPE}. All of these tasks are at least NP-hard in general, and therefore approximate methods are often preferred over exact ones in practice.}

\sva{We focus on three tasks over MRFs: (1) structure learning; (2) posterior marginal inference; and (3) finding the most likely assignment to all the non-evidence variables given evidence (this task is often called most probable explanation or MPE inference in short). All of these tasks are at least NP-hard in general, and therefore approximate methods are often preferred over exact ones in practice.}

A popular and fast method for structure learning is to learn binary pairwise MRFs (MRFs in which each feature is defined over at most two variables) by training an $\ell_1$-regularized logistic regression classifier for each variable given all other variables as features \cite{NIPS2006_Wainwright_86b20716,NIPS2006_lee_a4380923}. $\ell_1$-regularization induces sparsity in that it encourages many weights to take the value zero. All non-zero weights are then converted into conjunctive features. Each conjunctive feature evaluates to $\texttt{True}$ if both variables are assigned the value $1$ and to $\texttt{False}$ otherwise. Popular approaches for posterior marginal inference are the Gibbs sampling algorithm and generalized Belief propagation \cite{yedidiaGeneralizedBeliefPropagation2000} techniques such as Iterative Join Graph Propagation \cite{mateescu2010join}. For MPE inference, a popular approach is to encode it as an integer linear programming (ILP) problem \cite{koller2009probabilistic} and then use off-the-shelf approaches such as \citet{gurobi} to solve the ILP.

\textbf{Dependency Networks (DNs)} \cite{heckerman2000dependency} represent the joint distribution using a set of local conditional probability distributions, one for each variable. Each conditional distribution defines the probability of a variable given all of the others. A DN is consistent if there exists a joint probability distribution $P(\rvx)$ such that all conditional distributions $P_i(x_i|\rvx_{-i})$ where $\rvx_{-i}$ is the projection of $\rvx$ on $\rmX \setminus 
\{X_i\}$, are conditional distributions of $P(\rvx)$. 

A DN is learned from data by learning a  classifier (e.g., logistic regression, multi-layer perceptron, etc.) for each variable, and thus DN learning is embarrassingly parallel. However, because the classifiers are independently learned from data, we often get an inconsistent DN. It has been conjectured \cite{heckerman2000dependency} that most DNs learned from data are almost consistent in that only a few parameters need to be changed in order to make them consistent.

The most popular inference method over DNs is \textit{fixed-order} Gibbs sampling \cite{liubook}. If the DN is consistent, then its conditional distributions are derived from a
joint distribution $P(\rvx)$, and the stationary distribution (namely the distribution that Gibbs sampling converges to) will be the same as $P(\rvx)$. If the DN is inconsistent, then the
stationary distribution of Gibbs sampling will be inconsistent with the conditional distributions.

%% file: content/ddn.tex
\old{In this section, we describe how to solve the multi-label action classification task in videos and the multi-label image classification task using a hybrid of dependency networks and neural networks. At a high level, the neural network provides high-quality features given video segments/images and the dependency network represents and reasons about the relationships between the labels and features.}

In this section, we detail the training process for our proposed Deep Dependency Network model \cite{guo_2020_deep}, the hybrid framework for multi-label action classification in videos and multi-label image classification. Two key components are trained jointly: a neural network and a conditional dependency network. The neural network is responsible for extracting high-quality features from video segments or images, while the dependency network models the relationships between these features and their corresponding labels, supplying the neural network with gradient information regarding these relationships. 

\eat{In this section, we describe how to solve the multi-label action classification task in videos using a hybrid of dependency networks and neural networks. At a high level, the neural network provides high-quality features given videos and the dependency network represents and reasons about the probabilistic relationships between the labels and features.}

\subsection{Framework}
\begin{figure}[t]
\begin{center}
\centerline{\includegraphics[width=.9\linewidth]{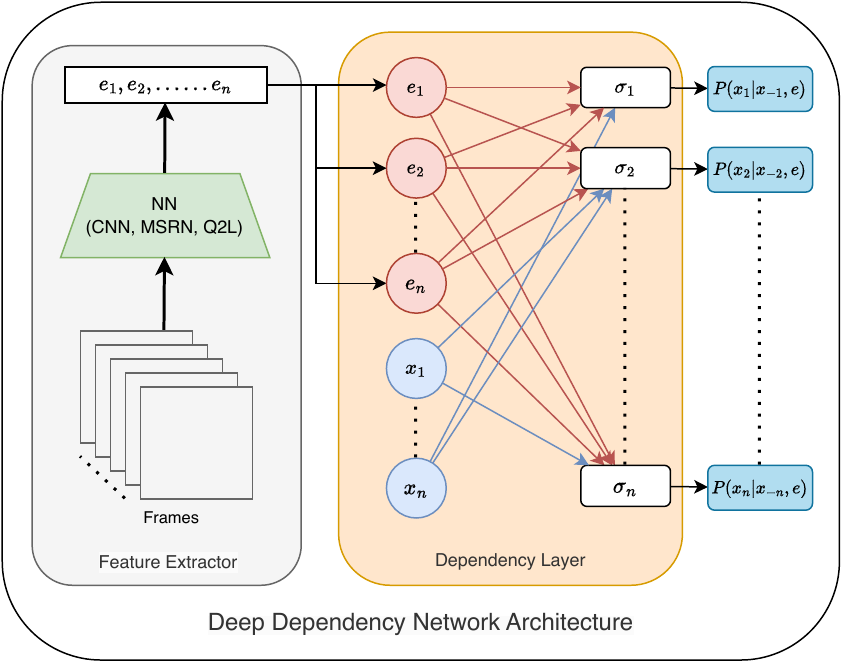}}
        \caption{Illustration of Dependency Network for multi-label video classification. The NN takes video clips (frames) as input and outputs the features $e_1,e_2,...,e_n$ (denoted by red colored nodes). These features are then used by the sigmoid output ($\sigma_1$, $\ldots$, $\sigma_n$) of the dependency layer to model the local conditional distributions. \eat{At each output node (blue boxes), the form of the conditional distribution is variable given its parents (incoming arrows represented by orange and blue color).}}
        \label{fig:ddn}
        \end{center}
\vspace{-8mm}
\end{figure}
Let $\rmV$ denote the set of random variables corresponding to the pixels and $\rvv$ denote the RGB values of the pixels in a frame or a video segment. Let $\rmE$ denote the (continuous) output nodes of a neural network which represents a function $\mathcal{N}: \rvv \mapsto \rve$, that takes $\rvv$ as input and outputs an assignment $\rve$ to $\rmE$. Let $\rmX=\{X_1,\ldots,X_n\}$ denote the set of indicator variables representing the labels. For simplicity, we assume that $|\rmE|=|\rmX|=n$. Given $(\rmV,\rmE,\rmX)$, a deep dependency network (DDN) is a pair $ \langle \mathcal{N},\mathcal{D} \rangle$ where $\mathcal{N}$ is a neural network that maps $\rmV=\rvv$ to $\rmE=\rve$ and $\mathcal{D}$ is a conditional dependency network \cite{guo2011multi} that models 
$P(\rvx|\rve)$ where $\rve=\mathcal{N}(\rvv)$. The conditional dependency network represents the distribution $P(\rvx|\rve)$ using a collection of local conditional distributions $P_i(x_i|\rvx_{-i},\rve)$, one for each label $X_i$, where $\rvx_{-i} = \{ x_1, \ldots, x_{i-1}, x_{i+1}, \ldots, x_{n}\}$. 

Thus, a DDN is a discriminative model and represents the conditional distribution  $P(\rvx|\rvv)$ using several local conditional distributions $P(x_i|\rvx_{-i},\rve)$ and makes the following conditional independence assumptions $P(x_i|\rvx_{-i},\rvv)=P(x_i|\rvx_{-i},\rve)$ where $\rve=\mathcal{N}(\rvv)$. Figure~\ref{fig:ddn} demonstrates the DDN architecture.
\subsection{Learning}
\old{We can either train the DDN using a \textit{pipeline method} or via \textit{joint training}. In the pipeline method, we first train the neural network using standard approaches (e.g., using cross-entropy loss) or use a pre-trained model. Then for each training example $(\rvv,\rvx)$, we send the video/image through the neural network to obtain a new representation $\rve$ of $\rvv$. The aforementioned process transforms each training example $(\rvv,\rvx)$ into a new feature representation $(\rve,\rvx)$ where $\rve=\mathcal{N}(\rvv)$. Finally, for each label $X_i$, we learn a classifier to model the conditional distribution $P_i(x_i|\rvx_{-i},\rve)$. Specifically, given a training example $(\rve,\rvx)$, each sigmoid output of the dependency layer indexed by $i$ ($\sigma_i$), uses $X_i$ as the class variable and $(\rmE \cup \rmX_{-i})$ as the attributes.

The pipeline method has several useful properties: it requires modest computational resources, is relatively fast and can be easily parallelized. As a result, it is especially beneficial when (only) less powerful GPUs are available at training time but a pre-trained network that is trained using more powerful GPUs is readily available.}

\sva{We employ the conditional pseudo log-likelihood loss (CPLL) \cite{besag_statistical_1975} in order to jointly train the two components of DDN, drawing inspiration from the DDN training approach outlined by \citet{guo_2020_deep}.
For each training example $(\rvv,\rvx)$, we send the video/image through the neural network to obtain a new representation $\rve$ of $\rvv$. In the dependency layer, we learn a classifier for each label $X_i$ to model the conditional distribution $P_i(x_i|\rvx_{-i},\rve)$. More precisely, with the representation of the training instance $(\rve,\rvx)$, each sigmoid output of the dependency layer indexed by $i$  and denoted by $\sigma_i$ (see figure~\ref{fig:ddn}) uses $X_i$ as the class variable and $(\rmE \cup \rmX_{-i})$ as the attributes. The joint training of the model is achieved through CPLL applied to the outputs of the dependency layer ($\sigma_i$'s).}


\eat{\vibhav{Need to change the stuff below. It does not sound right. The loss as presented makes no intuitive sense. 
To perform \textit{joint training}, we propose to use negative conditional pseudo log-likelihood \cite{besag_statistical_1975} as our loss function. 

We can say something like; easiest loss to use is cross entropy at the dependency net layer. However, its performance was not good. The issue is that it will get rid of the original intended meaning of the output nodes in the neural network; the goal of each node was also to predict the action. Therefore, just at the output node, we also add the original output loss. 
}}

\eat{Within the joint learning framework, we employ the conditional pseudo log-likelihood loss (CPLL) \cite{besag_statistical_1975}, drawing inspiration from the DDN training approach outlined by \cite{guo_2020_deep}.} 

Let $\Theta$ represent the parameter set of the DDN. We employ gradient-based optimization methods (e.g., backpropagation), to minimize the Conditional Pseudo-Likelihood (CPLL) loss function given below
\eat{
\begin{align}
\mathcal{L}(\Theta,\rvv,\rvx)&= - \sum_{i=1}^{n} \log P_i(x_i|\rvv,\rvx_{-i}; \Theta) \\
\label{eq:loss1}
&=- \sum_{i=1}^{n}\log P_i(x_i|\rve=\mathcal{N}(\rvv),\rvx_{-i}; \Theta)
\end{align}
}
\begin{equation}
\mathcal{L}(\Theta,\rvv,\rvx) = - \sum_{i=1}^{n}\log P_i(x_i|\rve=\mathcal{N}(\rvv),\rvx_{-i}; \Theta)
\label{eq:loss1}
\end{equation}
\old{In practice, for faster training/convergence, we will partition the parameters $\Theta$ of the DDN into two (disjoint) subsets $\Pi$ and $\Gamma$ where $\Pi$  and $\Gamma$ denote the parameters of the neural network and local conditional distributions respectively; and initialize $\Pi$ using a pre-trained neural network. Then, we can use any gradient-based (backpropagation) method to minimize the loss function.}

%% file: content/inference.tex
Unlike a conventional discriminative model, such as a neural network, in a DDN, we cannot predict the output labels by simply making a forward pass over the network. This is because each sigmoid output $\sigma_i$ (which yields a probability distribution over $X_i$) of the dependency layer requires an assignment $\rvx_{-i}$ to all labels except $x_i$ and $\rvx_{-i}$ is not available at prediction time. Consequently, it is imperative to employ specialized techniques for obtaining output labels in multilabel classification tasks within the context of DDNs.

Given a DDN representing a distribution $P(\rvx|\rve)$ where $\rve=\mathcal{N}(\rvv)$, the multilabel classification task can be solved by finding the most likely assignment to all the unobserved variables based on a set of observed variables. This task is also called the most probable explanation (MPE) task. Formally, we seek to find $\rvx^*$ such that:
\begin{equation}
\rvx^* = \argmax_\rvx P(\rvx \mid \rve)    
\end{equation}
Next, we present three algorithms for solving the MPE task.


\subsection{Gibbs Sampling}
\label{sec:gibbs}
To perform MPE inference, we first send the video (or frame) through the neural network to yield an assignment $\rve$ to all variables in $\rmE$. Then given $\rve$, we generate $N$ samples $(\rvx^{(1)},\ldots,\rvx^{(N)})$ via Gibbs sampling (see for example \citet{koller2009probabilistic}), a classic MCMC technique. These samples can then be used to estimate the marginal probability distribution of each label $X_i$ using the following mixture estimator \cite{liubook}:
\begin{align}
\label{eq:Mixture_Estimator1}
\hat{P}_i\left(x_{i}|\rvv\right)=\frac{1}{N} \sum_{j=1}^{N} P_i\left(x_{i} \mid \rvx_{-i}^{(j)},\rve\right)
\end{align}
Given an estimate of the marginal distribution $\hat{P}_i\left(x_{i}|\rvv\right)$ for each variable $X_i$, we can estimate the MPE assignment using the standard max-marginal approximation:
\[ \max_{\rvx} P(\rvx|\rve) \approx \prod_{i=1}^{n}\max_{x_i} \hat{P}(x_i|\rve) \]
In other words, we can construct an approximate MPE assignment by finding the value $x_i$ for each variable that maximizes $\hat{P}_i\left(x_{i}|\rve\right)$.

\eat{
The inference in DNs has been conducted using a Gibbs sampling-based method, as elaborated in the appendix \citep{heckerman2000dependency}. After obtaining $\mathbf{e}$, we perform fixed-order Gibbs Sampling over the dependency network (DN) where the latter represents the distribution $P(\rvx|\rve)$. Finally, given samples $(\rvx^{(1)},\ldots,\rvx^{(N)})$ generated via Gibbs sampling, we estimate the marginal probability distribution of each label $X_i$ using the following mixture estimator \cite{liubook}:
\begin{align}
\label{eq:Mixture_Estimator1}
\hat{P}_i\left(x_{i}|\rvv\right)=\frac{1}{N} \sum_{j=1}^{N} P_i\left(x_{i} \mid \rvx_{-i}^{(j)},\rve\right)
\end{align}

\sva{Gibbs Sampling is a method that provides an estimate of Max Marginals, functioning as a possible approximation to the MPE. This technique generates the marginal distribution for each variable, expressed as $\forall i \max_{x_{i}} P_{i}\left(x_{i}|\mathbf{v}\right)$.} 
}
\subsection{Local Search Based Methods}

Local search algorithms (see for example \citet{Selman1993LocalSS}) systematically examine the solution space by making localized adjustments, with the objective of either finding an optimal solution or exhausting a pre-established time limit. They offer a viable approach for solving the MPE inference task in DDNs. These algorithms, through their structured exploration and score maximization, are effective in identifying near-optimal label configurations.

In addressing the MPE inference within DDNs, we define the objective function for local search as $\sum_{i=1}^{n} \log \left( P_i(x_{i} \mid \rvx_{-i},\rve)\right)$, where $\rve$ is the evidence provided by $\mathcal{N}$. We propose to use two distinct local search strategies for computing the MPE assignment: random walk and greedy local search.  

In \textit{random walk (RW)}, the algorithm begins with a random assignment to the labels and at each iteration, flips a random label (changes the value of the label from a 1 to a 0 or a 0 to a 1) to yield a new assignment. At termination, the algorithm returns the assignment with the highest score explored during the random walk. In \textit{greedy local search}, the algorithm begins with a random assignment to the labels, and at each iteration, with probability $p$ flips a random label to yield a new assignment and with probability $1-p$ flips a label that yields the maximum improvement in the score. At termination, the algorithm returns the assignment with the highest score explored during its execution.
\subsection{Multi-Linear Integer Programming}
In this section, we present a novel approach for MPE inference in DDNs by formulating the problem as a Second-Order Multi-Linear Integer Programming task. Specifically, we show that the task of maximizing the scoring function $\sum_{i=1}^{n} \log \left( P_i(x_{i} \mid \rvx_{-i},\rve)\right)$ is equivalent to the following optimization problem (derivation is provided in the supplementary material):
\begin{equation}
\vspace{-5pt}
\label{eq:opt_ilp}
\begin{aligned}
& \underset{\rvx, \rvz}{\text{maximize}} \sum_{i=1}^n \left( x_i \log P_i + (1 - x_i) \log(1 - P_i) \right) \\
& \text{subject to:} \\
& \quad P_i = \sigma(z_i), \quad \forall i \in \{1,\ldots,n\} \\
& \quad z_i = \sum_{j=1}^{|\rve|}w_{ij} e_j + \sum_{\substack{k=1 \\ k \neq i}}^{|\rvx|} v_{ik} x_k + b_i, \quad \forall i \in \{1,\ldots,n\} \\
& \quad x_i \in \{0, 1\}, \quad \forall i \in \{1,\ldots,n\}
\end{aligned}
\end{equation}

Here, $P_i = P_i(x_i | \rvx_{-i},\rve)$ and $w_{ij}$'s and $v_{ik}$'s denote the weights associated with $\rve$ and $\rvx$ for $P_i$, respectively. The bias term for each $P_i$ is denoted by $b_i$. In this context, $z_i$ represents the values acquired prior to applying the sigmoid activation function. Substituting the constraint $P_i(x_i | \rvx_{-i},\rve) = \sigma(z_i) = \frac{1}{1 + e^{(- z_{i})}}$ in the objective and simplifying, we get
\eat{By utilizing the formulation of each sigmoid output $\sigma_i$, the formulation for $P_i$ is provided below:
\begin{equation}
\label{eq:opt_logits}
P_i(x_i | \rvx_{-i},\rve) = \frac{1}{1 + e^{(- \rvz_{i})}}
\end{equation}}
\begin{align}
&\underset{\rvx, \rvz}{\text{maximize}} \sum_{i=1}^n x_i z_{i} - \log \left(1 + e^{z_{i}}\right) \label{eq:opt_ilp_obj} \\
&\text{subject to:} \nonumber \\
&z_i = \sum_{j=1}^{|\rve|}w_{ij} e_j + \sum_{\substack{k=1 \\ k \neq i}}^{|\rvx|} v_{ik} x_k + b_i, \quad \forall i \in \{1,\ldots,n\} \label{eq:opt_ilp_z} \\
&x_i \in \{0, 1\}, \quad \forall i \in \{1,\ldots,n\} \label{eq:opt_ilp_x}
\end{align}

The optimal value for $\rvx$ corresponds to the solution of this optimization problem. The second term in the objective specified in \eqref{eq:opt_ilp_obj} comprises logarithmic and exponential functions, which are non-linear, thereby making it a \textit{non-linear optimization problem}. \sva{To address this non-linearity, we propose the use of piece-wise linear approximations}\citep{lin_2013_review, geissler_2012_using, kumar__converting, li_2022_fpga, asghari_2022_transformation, rovatti_2014_optimistic} for these terms. \sva{Further details about the piece-wise linear approximation can be found in the supplement.} Let $g(z_i)$ represent the piece-wise linear approximation of $\log \left(1 + e^{z_{i}}\right)$, then the optimization problem can be expressed as:
\begin{equation}
\label{eq:opt_ilp_small}
\begin{aligned}
&\underset{\rvx, \rvz}{\text{maximize}} \sum_{i=1}^n x_i z_{i} - g(z_i) \\
&\text{subject to:} \\
&z_i = \sum_{j=1}^{|\rve|}w_{ij} e_j + \sum_{\substack{k=1 \\ k \neq i}}^{|\rvx|} v_{ik} x_k + b_i, \quad \forall i \in \{1,\ldots,n\} \\
&x_i \in \{0, 1\}, \quad \forall i \in \{1,\ldots,n\}
\end{aligned}
\end{equation}

The optimization problem given in \eqref{eq:opt_ilp_small} is an integer multilinear program of order $2$ because it includes terms of the form $x_ix_j$ where both $x_i$ and $x_j$ take values from the set $\{0,1\}$. Since $x_ix_j$ corresponds to a ``logical and'' between two Boolean variables, all such expressions can be easily encoded as linear constraints yielding an integer linear program (ILP) (see the supplementary material for an example). 

In our experiments, we solved the ILP using Gurobi \cite{gurobi}, which is an anytime ILP solver. Another useful feature of the ILP formulation is that we can easily incorporate prior knowledge (e.g., an image may not have more than ten objects/labels) into the ILP under the assumption that the knowledge can be reliably modeled using linear constraints.


%% file: content/6-experiment.tex
In this section, we evaluate the proposed methods on two multi-label classification tasks: (1) multi-label activity classification using three video datasets; and (2) multi-label image classification using three image datasets. We begin by describing the datasets and metrics, followed by the experimental setup, and conclude with the results. All models were implemented utilizing PyTorch and were trained and evaluated on a machine with an NVIDIA A40 GPU and an Intel(R) Xeon(R) Silver 4314 CPU.
\subsection{Datasets and Metrics}
\label{sec:datasets}
\old{
We evaluated our algorithms on the following three video datasets: (1) Charades \cite{charades_paper}; (2) Textually Annotated Cooking Scenes (TACoS) \cite{tacos:regnerietal:tacl}; and (3) Wetlab \cite{naim-etal-2015-discriminative}. In the Charades dataset, the videos are divided into segments (video clips), and each segment is annotated with one or more action labels. In the TACoS and Wetlab datasets, each frame is associated with one or more actions. 

\textbf{Charades dataset} \cite{charades_paper} comprises videos of people performing daily indoor activities while interacting with various objects.\eat{It is a multi-label activity classification dataset; thus, more than one activity can happen at a given time.} In the standard split, there are 7,986 training videos and 1,863 validation videos\eat{, averaging 30 seconds}. We used the training videos to train the models and the validation videos for testing purposes. We follow the instructions provided in PySlowFast \cite{fan2020pyslowfast} to do the train-test split for the dataset. The dataset has roughly 66,500 temporal annotations for 157 action classes. \eat{We report mean Average Precision (mAP) following the standard evaluation protocols for this dataset. We also report Label Ranking Average Precision (LRAP) and Jaccard Index (JI), as these metrics have been used extensively in previous work for evaluating the performance of multi-label classifiers.} 



The \textbf{\eat{The Textually Annotated Cooking Scenes (}TaCOS dataset} \cite{tacos:regnerietal:tacl} consists of third-person videos of a person cooking in a kitchen. The dataset comes with hand-annotated labels of actions, objects, and locations for each video frame. From the complete set of these labels, we selected 28 labels. 
\eat{Each of these labels corresponds to a location, an object, or a verb, and each action is defined as a (location, object, verb) triple. By selecting the videos that correspond to these labels and}
By dividing the videos corresponding to these labels into train and test sets, we get a total of 60,313 frames for training and 9,355 frames for testing, spread out over 17 videos. 

The \textbf{Wetlab dataset} \cite{wetlab:Naim2014UnsupervisedAO} comprises videos where experiments are being performed by lab technicians that involve hazardous chemicals in a wet laboratory. 
We used five videos for training and one video for testing. The training set comprises 100,054 frames, and the test set comprises 11,743 frames.\eat{Each activity in the dataset can have multiple objects per activity (for example, an activity ``transfer contents'' can have two objects linked with it, a test tube and a beaker), and there are 57 possible labels.} There are 57 possible labels and each label corresponds to an object or a verb, and each action is made of one or more labels from each category.

We also evaluated our algorithms on three multi-label image classification (MLIC) datasets: (1) MS-COCO \cite{linMicrosoftCOCOCommon2015}; (2) PASCAL VOC 2007,  and  \cite{everinghamPascalVisualObject2010};  and (3) NUS-WIDE \cite{chuaNUSWIDERealworldWeb2009a}.

\textbf{MS-COCO} (Microsoft Common Objects in Context) \cite{linMicrosoftCOCOCommon2015} is a large-scale object detection and segmentation dataset. It has also been extensively used for the MLIC task. The dataset contains 122,218 labeled images and 80 labels in total. Each image is labeled with at least 2.9 labels on average. We used the 2014 version of the dataset\eat{ as mentioned in \cite{liuQuery2LabelSimpleTransformer2021}}.

\textbf{NUS-WIDE} dataset \cite{chuaNUSWIDERealworldWeb2009a} is a real-world web image dataset that contains 269,648 images from Flickr. Each image has been manually annotated with a subset of 81 visual classes that include objects and scenes. 

\textbf{PASCAL VOC 2007} \citep{everinghamPascalVisualObject2010} is another dataset that has been used widely for the MLIC task.
The dataset contains 5,011 images in the train-validation set and 4,952 images in the test set. The total number of labels in the dataset is 20 which corresponds to object classes.\eat{ and each image can take one or more labels. }
}

\sva{We evaluated our algorithms on three video datasets: (1) Charades \cite{charades_paper}; (2) TACoS \cite{tacos:regnerietal:tacl}; and (3) Wetlab \cite{naim-etal-2015-discriminative}. Charades dataset comprises videos of people performing daily indoor activities while interacting with various objects. We adopted the train-test split instructions given in  PySlowFast \cite{fan2020pyslowfast} with 7,986 training and 1,863 validation videos. TACoS consists of third-person videos of a person cooking in a kitchen. The dataset comes with hand-annotated labels of actions, objects, and locations for each video frame. From the complete set of these labels, we selected 28 labels resulting in 60,313 training frames and 9,355 test frames across 17 videos. Wetlab features experimental activities in labs, consisting of five training videos (100,054 frames) and one test video (11,743 frames) with 57 distinct labels.

For multi-label image classification (MLIC), we examined: (1) MS-COCO \cite{linMicrosoftCOCOCommon2015}; (2) PASCAL VOC 2007 \cite{everinghamPascalVisualObject2010}; and (3) NUS-WIDE \cite{chuaNUSWIDERealworldWeb2009a}. MS-COCO, a well-known dataset for detection and segmentation, comprises 122,218 labeled images with an average of 2.9 labels per image. We used its 2014 version. NUS-WIDE is a real-world web image dataset that contains 269,648 images from Flickr. Each image has been manually annotated with a subset of 81 visual classes that include objects and scenes. PASCAL VOC 2007 contains 5,011 train-validation and 4,952 test images, with each image labeled with one or more of the 20 available object classes.}


\old{We follow the instructions provided in \cite{quMultilayeredSemanticRepresentation2021} to do the train-test split for NUS-WIDE and PASCAL VOC 2007. 
We evaluated the performance on the TACoS, Wetlab, MS-COCO, NUS-WIDE, and VOC datasets using the following four metrics: mean Average Precision (mAP), Label Ranking Average Precision (LRAP), Subset Accuracy (SA), and Jaccard Index (JI). For all the metrics that are being considered here, a higher value means better performance. mAP and LRAP require access to an estimate of the posterior probability distribution at each label while SA and JI are non-probabilistic and only require an assignment to the labels. 
Note that we only report mAP, LRAP, and JI on the Charades dataset because existing approaches cannot achieve reasonable SA due to the large size of the label space.

In recent years, mAP has been used as an evaluation metric for multi-label image and action classification in lieu of conventional metrics such as SA and JI. However, both SA (which seeks an exact match with the ground truth) and JI are critical for applications such as dialogue systems \cite{vilarMultilabelTextClassification2004}, self-driving cars \cite{mlfselfdriving, protopapadakisMultilabelDeepLearning2020} and disease diagnosis \cite{maxwellDeepLearningArchitectures2017, zhangNovelDeepNeural2019, zhouApplicationMultilabelClassification2021} where MLC is one of the sub-tasks in a series of interrelated sub-tasks. Missing a single label in these applications could have disastrous consequences for downstream sub-tasks.
}

\sva{We follow the instructions provided in \cite{quMultilayeredSemanticRepresentation2021} to do the train-test split for NUS-WIDE and PASCAL VOC. We evaluated the performance on the TACoS, Wetlab, MS-COCO, NUS-WIDE, and VOC datasets using Subset Accuracy (SA), Jaccard Index (JI), Hamming Loss (HL), Macro F1 Score (Macro F1), Micro F1 Score (Micro F1) and F1 Score (F1). With the exception of Hamming Loss, superior performance is indicated by higher scores in all considered metrics. We omit SA for the Charades dataset due to the infeasibility of achieving reasonable scores given its large label space. Additionally, Hamming Loss (HL) has been excluded for the MS-COCO dataset as the performance of all methods is indistinguishable.

Given that the focus in MPE inference is on identifying the most probable label configurations rather than individual label probabilities, the use of Mean Average Precision (mAP) as a performance metric is not applicable to our study. Therefore, our primary analysis relies on SA, JI, HL, Macro F1, Micro F1, and F1. Precision-based metrics are detailed in the supplementary material.}

\subsection{Experimental Setup and Methods}
We used three types of architectures in our experiments: (1) Baseline neural networks, which are specific to each dataset; (2) neural networks augmented with MRFs, which we will refer to as deep random fields or DRFs in short; and (3) a dependency network on top of the \icml{neural networks} called deep dependency networks (DDNs). 

\textbf{Neural Networks}. \icml{We choose four different types of neural networks, and they act as a baseline for the experiments and as a feature extractor for DRFs and DDNs. Specifically, we experimented with: (1) 2D CNNs, (2) 3D CNNs, (3) transformers, and (4) CNNs with attention module and graph attention networks (GAT) \cite{velivckovicgraph}. 
This helps us show that our proposed method can improve the performance of a wide variety of neural architectures, even those which model label relationships, because unlike the latter, it performs probabilistic inference.}

For the Charades dataset, we use the PySlowFast \cite{fan2020pyslowfast} implementation of the SlowFast Network \cite{Feichtenhofer_2019_ICCV} (a state-of-the-art 3D CNN for video classification). For TACoS and Wetlab datasets, we use InceptionV3 \cite{szegedyRethinkingInceptionArchitecture2016}, a state-of-the-art 2D CNN model for image classification. \icml{For the MS-COCO dataset, we used Query2Label (Q2L) \cite{liuQuery2LabelSimpleTransformer2021}, which uses transformers to pool class-related features. Q2L also learns label embeddings from data to capture the relationships between the labels. 
Finally, we used the multi-layered semantic representation network (MSRN) \cite{quMultilayeredSemanticRepresentation2021} for NUS-WIDE and PASCAL VOC. MSRN also models label correlations and learns semantic representations at multiple convolutional layers. We adopt pre-trained models and hyper-parameters from existing repositories for Charades, MS-COCO, NUS-WIDE, and PASCAL VOC.} For TaCOS and Wetlab datasets, we fine-tuned an InceptionV3 model that was pre-trained on the ImageNet dataset. 

\textbf{Deep Random Fields (DRFs)}. As a baseline, we used a model that combines MRFs with \icml{neural networks}. This DRF model is similar to DDN except that we use an MRF instead of a DN to compute $P(\rvx|\rve)$. We trained the MRFs generatively; namely, we learned a joint distribution $P(\rvx,\rve)$, which can be used to compute $P(\rvx|\rve)$ by instantiating evidence. We chose generative learning because we learned the structure of the MRFs from data, and discriminative structure learning is slow in practice \cite{koller2009probabilistic}. 

\eat{Specifically, we used the logistic regression with $\ell_1$ regularization method of \cite{NIPS2006_Wainwright_86b20716} to learn a pairwise MRF. The training data for this method is obtained by sending each annotated video clip (or frame) $(\rvv,\rvx)$ through the neural network and transforming it to $(\rve,\rvx)$ where $\rve=\mathbb{N}(\rvv)$. At termination, this method yields a graph $\gG$ defined over $\rmX \cup \rmE$. 

For parameter/weight learning, we converted each edge over $\rmX \cup \rmE$ to a conjunctive feature. For example, if the method learns an edge between $X_i$ and $E_j$, we use a conjunctive feature $X_i \wedge E_j$ which is true if both $X_i$ and $E_j$ are assigned the value $1$. Then we learned the weights for each feature by maximizing the pseudo log-likelihood of the data.
}

For inference over MRFs, we used Gibbs sampling (GS), Iterative Join Graph Propagation (IJGP) \cite{mateescu2010join}, and Integer Linear Programming (ILP) methods. Thus, three versions of DRFs corresponding to the inference scheme were used. We refer to these schemes as DRF-GS, DRF-ILP, and DRF-IJGP, respectively. Note that IJGP and ILP are advanced schemes, and we are unaware of their use for \icml{multi-label  classification}.\eat{multi-label action classification \icml{and multi-label image classification}}
Our goal is to test whether advanced inference schemes help improve the performance of deep random fields.

\old{\textbf{Deep Dependency Networks (DDNs)}. We experimented with four versions of DDNs: (1) DDN-LR-Pipeline; (2) DDN-MLP-Pipeline; (3) DDN-LR-Joint; and (4) DDN-MLP-Joint. The first and third versions use logistic regression (LR), while the second and fourth versions use multi-layer perceptrons (MLP) to represent the conditional distributions. The first two versions are trained using the pipeline method, while the last two versions are trained using the joint learning loss given in \eqref{eq:loss1}.
}

\textbf{Deep Dependency Networks (DDNs)}. We trained the Deep Dependency Networks (DDNs) using the joint learning loss described in \eqref{eq:loss1}. We examined four unique inference methods for DDNs: (1) DDN-GS, employing Gibbs Sampling; (2) DDN-RW, leveraging a random walk local search; (3) DDN-Greedy, implementing a greedy local search; and (4) DDN-ILP, utilizing Integer Linear Programming to optimize the objective given in \eqref{eq:opt_ilp_small}. 
It is noteworthy that, until now, only DDN-GS has been used for inference in Dependency Networks. DDN-RW and DDN-Greedy, which are general-purpose local search techniques, are our proposals for MPE inference on Dependency Networks. Lastly, DDN-ILP introduces a novel approach using optimization techniques with the objective of enhancing MPE inference on dependency networks.

\textbf{Hyperparameters.} For DRFs, in order to learn a sparse structure (using the logistic regression with $\ell_1$ regularization method of \citet{NIPS2006_Wainwright_86b20716}), we increased the regularization constant associated with the $\ell_1$ regularization term until the number of neighbors of each node in $\gG$ is bounded between 2 and 10. We enforced this sparsity constraint in order to ensure that the inference schemes, specifically IJGP and ILP, are accurate and the model does not overfit the training data. IJGP, ILP, and GS are anytime methods; for them, we used a time-bound of one minute per example. 

\old{For DDNs, we used LR with $\ell_1$ regularization and MLPs with $\ell_2$ regularization. For MLP the number of hidden layers was selected from the $\{2,3,4\}$. The regularization constants for LR and MLP (chosen from the $\{0.1,0.01,0.001\}$) and the number of layers for MLP were chosen via cross-validation. For all datasets, MLP with four layers performed the best. ReLU was used for the activation function for each hidden layer and sigmoid for the outputs. For joint learning, we reduced the learning rates of both LR and MLP models by expanding on the learning rate scheduler given in PySlowFast \cite{fan2020pyslowfast}\eat{. We used the steps with relative learning rate policy, where the learning rate depended on the current epoch} and the initial learning rate was chosen in the range of $\displaystyle [10^{-3}, 10^{-5}]$.}

\sva{For DDNs, we used $\ell_1$ regularization for the dependency layer. Through cross-validation, we selected the regularization constants from the set $\{0.1,0.01,0.001\}$. In the context of joint learning, the learning rates of the DDN model were adjusted using an extended version of the learning rate scheduler from PySlowFast \cite{fan2020pyslowfast}. We impose a strict time constraint of 60 seconds per example for DDN-GS, DDN-RW, DDN-Greedy, and DDN-ILP. The DDN-ILP method is solved using \citet{gurobi} and utilizes accurate piece-wise linear approximations for non-linear functions, subject to a pre-defined error tolerance of $0.001$. 
}


\eat{
need to have the number of neighbors each node can have. We select values in the range $\displaystyle [2,10]$ for these hyper-parameters. If we go beyond ten, then the inference part takes much time to complete. For the structure we provide the learning rates of $\displaystyle \{0.01, 0.001\}$ and the regularization parameters with the values $\displaystyle \{0.1, 0.01, 0.001\}$. For the inference techniques, we need to provide the maximum run time for the IJGP method, which we selected from $\displaystyle \{10, 30, 60\}$ seconds. We used Neural Networks with 3 and 4 hidden layers for the distributions of dependency networks. For all the models in dependency networks, we chose the learning rates from $\displaystyle \{0.05, 0.01, 0.001\}$ and regularization parameters from $\displaystyle \{0.1, 0.01, 0.001\}$. The number of samples for inference on the dependency network were from $\displaystyle \{1000, 10000, 50000\}$. For the joint learning experimentation, we reduced the learning rates of both models by expanding on the learning rate scheduler given with PySlowFast. We use the steps with relative learning rate policy, where the learning rate depended on the current epoch and was in the range of $\displaystyle [1e^{-3}, 1e^{-5}]$. Other hyper-parameters remain the same as the pipeline models. 
}
\subsection{Results}

\input{content/mlac_results_all_metrics}

\input{content/mlic_results_all_metrics}

We compare the baseline neural networks with three versions of DRFs and four versions of DDNs using the six metrics and six datasets given in Section \ref{sec:datasets}. The results are presented in tables \ref{tab:eval-mlac} and \ref{tab:eval-mlic}.


\old{
\textbf{Comparison between Baseline neural network and DRFs}. \icml{We observe that IJGP and ILP outperform the baseline neural networks (which includes transformers for some datasets) in terms of the two non-probabilistic metrics JI and SA on five out of the six datasets. IJGP typically outperforms GS and ILP on JI. ILP outperforms the baseline on SA (notice that SA is 1 if there is an exact match between predicted and true labels and 0 otherwise) because it performs an accurate maximum-a-posteriori (MAP) inference (accurate MAP inference on an accurate model is likely to yield high SA). 
However, on metrics that require estimating the posterior probabilities, mAP and LRAP, the DRF schemes sometimes hurt the performance and at other times are only marginally better than the baseline methods.
We observe that advanced inference schemes, particularly IJGP and ILP are superior on average to GS. Note that getting a higher SA is much harder in datasets having high label cardinalities. Specifically, SA does not distinguish between models that predict \textit{almost} correct labels and completely incorrect outputs.}

\textbf{Comparison between Baseline neural networks and DDNs}. We observe that the best performing DDN model, DDN-MLP with joint learning, outperforms the baseline neural networks on \icml{five out of the six datasets} on all metrics. Sometimes, the improvement is substantial (e.g., 9\% improvement in mAP on the wetlab dataset). \icml{The DDN-MLP model with joint learning improves considerably over the baseline method when performance is measured using SA and JI while keeping the precision comparative or higher.} Roughly speaking, the MLP versions are superior to the LR versions, and the joint models are superior to the ones trained using the pipeline method. On the non-probabilistic metrics (JI and SA), the pipeline models are often superior to the baseline neural network, while on the mAP metric, they may hurt the performance.
}

\textbf{Comparison between baseline neural network and DRFs}. We observe that IJGP and ILP outperform the baseline neural networks (which include transformers for some datasets) in terms of JI, SA, and HL on four out of the six datasets. IJGP typically outperforms GS and ILP on JI. In terms of F1 metrics, IJGP and ILP methods tend to perform better than Gibbs Sampling (GS) approaches, but they are less effective than baseline NNs. ILP's superiority in SA—a metric that scores 1 for an exact label match and 0 otherwise—can be attributed to its precise most probable explanation (MPE)  inference. An accurate MPE inference, when paired with a precise model, is likely to achieve high SA scores. Note that getting a higher SA is much harder in datasets having large number of labels. Specifically, SA does not distinguish between models that predict \textit{almost} correct labels and completely incorrect outputs. We observe that advanced inference schemes, particularly IJGP and ILP, are superior on average to GS. 

\textbf{Comparison between baseline neural networks and DDNs}. Our study has shown that the DDN model with the proposed MILP based inference method is superior to the baseline neural networks in four out of six datasets, with notable enhancements in SA and JI metrics, such as an 18\%, 23\%, and 30\% improvement in SA for the PASCAL-VOC, TACoS, and Wetlab datasets, respectively. Moreover, the MILP-based method either significantly surpasses or matches all alternative inference strategies for DDNs. While Gibbs Sampling-based inference is typically better than the Random Walk based approach in DDNs, its performance against the greedy sampling method varies.

We observe that on the MLIC task, the DDN with advanced MPE inference outperforms Q2L and MSRN, even though both Q2L and MSRN model label correlations. This suggests that DDNs are either able to uncover additional relationships between labels during the learning phase or better reason about them during the inference phase or both. In particular, both Q2L and MSRN do not use MPE inference to predict the labels because they do not explicitly model the joint probability distribution over the labels. 


\begin{figure*}[t!]
\begin{center}
\centerline{
  \includegraphics[width=\textwidth]{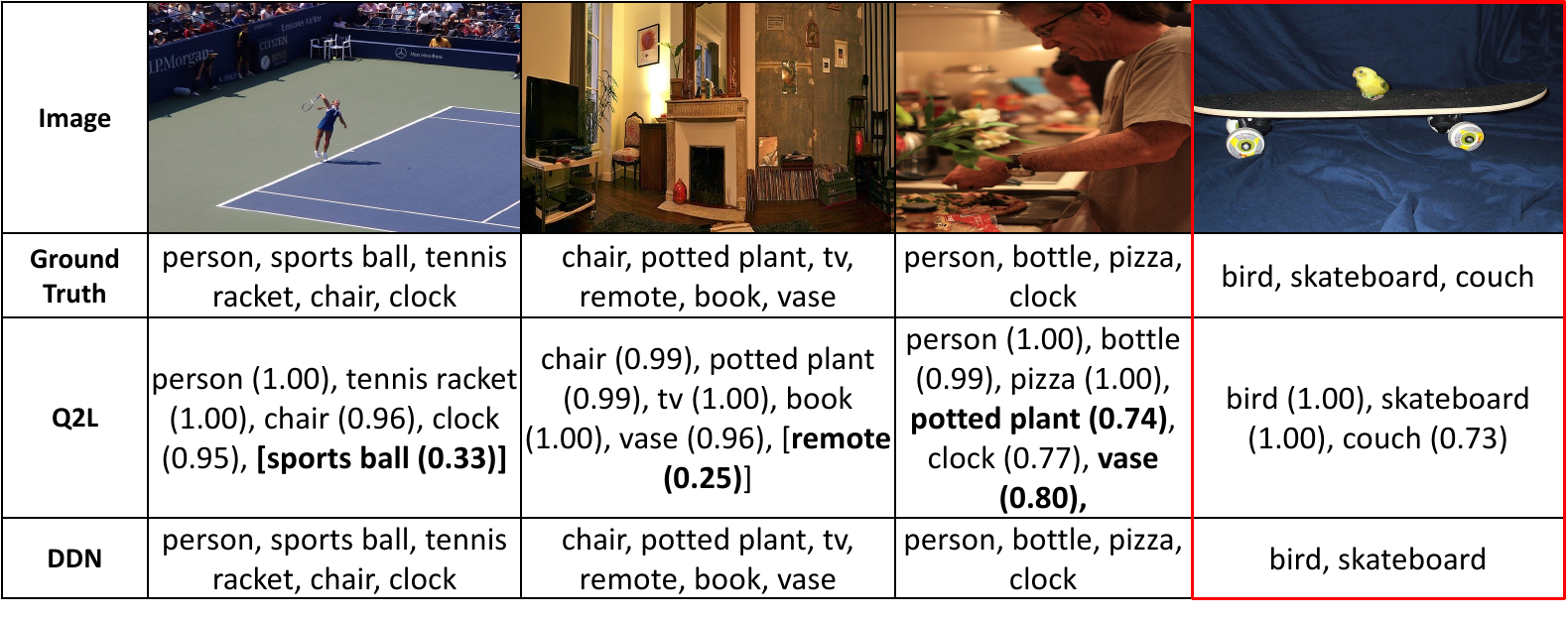}}
  \caption{Comparison of labels predicted by Q2L~\cite{liuQuery2LabelSimpleTransformer2021} and our DDN-ILP scheme on the MS-COCO dataset. Labels in bold represent the difference between the predictions of the two methods, assuming that a threshold of 0.5 is used (i.e., every label whose probability $ > 0.5$ is considered a predicted label). Due to the MPE focus in DDN-ILP, only label configurations are generated, omitting corresponding probabilities. The first three column shows examples where DDN improves over Q2L, while the last column (outlined in red) shows an example where DDN is worse than Q2L. }
  \label{tab:annotation_comparison}
  \end{center}
\vspace{-4mm}
\end{figure*}

\sva{Figure \ref{tab:annotation_comparison} displays the images and their predicted labels by both Q2L and DDN-ILP on the MS-COCO dataset. Our method not only adds the labels omitted by Q2L but also eliminates several incorrect predictions. In the first two images, our approach rectifies label omissions by Q2L, conforming to the ground truth. In the third image, our method removes erroneous predictions. The last image illustrates a case where DDN underperforms compared to Q2L by failing to identify a ground-truth label. Additional instances are available in the appendix.}

\sva{
\textbf{Comparison between DRFs and DDNs}. Based on our observations, we found that jointly trained DDNs in conjunction with the proposed inference method consistently lead to superior performance compared to the top-performing DRFs across all datasets. Nonetheless, in certain situations, DRFs that employ advanced inference strategies produce results that closely match those of DDNs for both JI and SA, making DRFs a viable option, particularly when limited GPU resources are available for training and optimization of both JI and SA is prioritized.

In summary, the empirical results indicate that Deep Dependency Networks (DDNs) equipped with advanced inference strategies consistently outperform conventional neural networks and MRF+NN hybrids in multi-label classification tasks. Furthermore, these advanced inference methods surpass traditional sampling-based techniques for DDNs. The superior performance of both DDNs and DRFs utilizing advanced inference techniques supports the value of such mechanisms and suggests that further advancement in this area has the potential to unlock additional capabilities within these models.}

%% file: content/mlac_results_all_metrics.tex
\begin{table*}[t]
\centering
\caption{Comparison of our methods with the feature extractor for MLAC. The best/second best values are bold/underlined.}
\label{tab:eval-mlac}
\resizebox{\textwidth}{!}{%
\begin{tabular}{cc|cc|ccc|cccc}
\hline
\textbf{Dataset} & \textbf{Metric} & \textbf{SlowFast} & \textbf{InceptionV3} & \textbf{DRF-GS} & \textbf{DRF-ILP} & \textbf{DRF-IJGP} & \textbf{DDN-GS} & \textbf{DDN-RW} & \textbf{DDN-Greedy} & \textbf{DDN-ILP} \\
\hline
\multirow{5}{*}{Charades} & JI & 0.29 & - & 0.22 & 0.31 & {\ul 0.32} & 0.31 & 0.30 & 0.31 & \textbf{0.33} \\
 & HL $\downarrow$ & \textbf{0.052} & - & 0.194 & 0.067 & {\ul 0.054} & {\ul 0.054} & 0.056 & 0.056 & \textbf{0.052} \\
 & Macro F1 & 0.32 & - & 0.16 & 0.18 & 0.19 & 0.30 & 0.13 & {\ul 0.33} & \textbf{0.36} \\
 & Micro F1 & {\ul 0.45} & - & 0.31 & 0.21 & 0.29 & 0.43 & 0.24 & 0.44 & \textbf{0.47} \\
 & F1 & {\ul 0.42} & - & 0.28 & 0.22 & 0.26 & 0.41 & 0.22 & 0.41 & \textbf{0.44} \\
\hline
\multirow{6}{*}{TACoS} & SA & - & 0.40 & 0.47 & 0.51 & 0.44 & 0.54 & 0.46 & {\ul 0.56} & \textbf{0.63} \\
 & JI & - & 0.61 & 0.65 & 0.65 & {\ul 0.70} & 0.69 & 0.64 & 0.69 & \textbf{0.72} \\
 & HL $\downarrow$ & - & 0.082 & 0.044 & \textbf{0.030} & 0.043 & 0.041 & 0.042 & {\ul 0.040} & {\ul 0.040} \\
 & Macro F1 & - & 0.26 & 0.54 & 0.42 & 0.54 & 0.52 & 0.59 & {\ul 0.61} & \textbf{0.62} \\
 & Micro F1 & - & 0.48 & 0.75 & 0.73 & \textbf{0.83} & 0.74 & 0.77 & {\ul 0.80} & 0.79 \\
 & F1 & - & 0.44 & 0.70 & 0.66 & \textbf{0.78} & 0.68 & 0.75 & 0.75 & {\ul 0.76} \\
\hline
\multirow{5}{*}{Wetlab} & SA & - & 0.35 & 0.35 & 0.60 & 0.58 & {\ul 0.63} & 0.46 & 0.55 & \textbf{0.65} \\
 & JI & - & 0.64 & 0.52 & 0.73 & 0.74 & \textbf{0.78} & 0.63 & 0.68 & {\ul 0.76} \\
 & HL $\downarrow$ & - & 0.017 & 0.058 & {\ul 0.014} & {\ul 0.014} & \textbf{0.011} & 0.025 & {\ul 0.014} & {\ul 0.014} \\
 & Macro F1 & - & 0.24 & 0.22 & {\ul 0.27} & 0.24 & \textbf{0.28} & 0.22 & \textbf{0.28} & \textbf{0.28} \\
 & Micro F1 & - & 0.76 & 0.69 & {\ul 0.81} & \textbf{0.82} & 0.80 & 0.66 & 0.80 & {\ul 0.81} \\
 & F1 & - & 0.72 & 0.68 & 0.77 & 0.77 & {\ul 0.79} & 0.68 & {\ul 0.79} & \textbf{0.80} \\
\hline
\end{tabular}%
}
\end{table*}

%% file: content/mlic_results_all_metrics.tex
\begin{table*}[t]
\centering
\caption{Comparison of our methods with the feature extractor for MLIC. The best/second best values are bold/underlined.}
\label{tab:eval-mlic}
\resizebox{\textwidth}{!}{%
\begin{tabular}{cc|cc|ccc|cccc}
\hline
\textbf{Dataset} & \textbf{Metric} & \textbf{Q2L} & \textbf{MSRN} & \textbf{DRF-GS} & \textbf{DRF-ILP} & \textbf{DRF-IJGP} & \textbf{DDN-GS} & \textbf{DDN-RW} & \textbf{DDN-Greedy} & \textbf{DDN-ILP} \\
\hline
\multirow{5}{*}{MS-COCO} & SA & 0.51 & - & 0.35 & {\ul 0.54} & \textbf{0.55} & \textbf{0.55} & 0.53 & \textbf{0.55} & \textbf{0.55} \\
 & JI & 0.80 & - & 0.69 & {\ul 0.82} & {\ul 0.82} & {\ul 0.82} & 0.81 & {\ul 0.82} & \textbf{0.83} \\
 & Macro F1 & \textbf{0.86} & - & 0.81 & 0.84 & {\ul 0.85} & \textbf{0.86} & 0.82 & {\ul 0.85} & {\ul 0.85} \\
 & Micro F1 & {\ul 0.86} & - & 0.82 & 0.83 & 0.85 & \textbf{0.87} & 0.85 & {\ul 0.86} & {\ul 0.86} \\
 & F1 & \textbf{0.88} & - & 0.79 & 0.82 & 0.85 & \textbf{0.88} & 0.86 & {\ul 0.87} & \textbf{0.88} \\
\hline
\multirow{6}{*}{NUS-WIDE} & SA & - & 0.31 & 0.28 & {\ul 0.32} & {\ul 0.32} & \textbf{0.33} & 0.25 & 0.30 & \textbf{0.33} \\
 & JI & - & {\ul 0.64} & 0.55 & 0.59 & 0.63 & 0.62 & 0.56 & 0.61 & \textbf{0.65} \\
 & HL $\downarrow$ & - & \textbf{0.015} & 0.020 & {\ul 0.016} & 0.017 & {\ul 0.016} & {\ul 0.016} & {\ul 0.016} & \textbf{0.015} \\
 & Macro F1 & - & \textbf{0.56} & 0.23 & 0.28 & 0.32 & {\ul 0.52} & 0.26 & 0.29 & 0.51 \\
 & Micro F1 & - & {\ul 0.73} & 0.63 & 0.70 & 0.71 & 0.71 & 0.70 & 0.72 & \textbf{0.74} \\
 & F1 & - & \textbf{0.71} & 0.62 & 0.67 & {\ul 0.69} & {\ul 0.69} & 0.68 & {\ul 0.69} & \textbf{0.71} \\
\hline
\multirow{5}{*}{PASCAL-VOC} & SA & - & 0.71 & 0.73 & 0.76 & 0.76 & 0.76 & 0.82 & {\ul 0.86} & \textbf{0.89} \\
& JI & - & 0.85 & 0.83 & 0.88 & 0.87 & 0.87 & 0.89 & {\ul 0.91} & \textbf{0.95} \\
 & HL $\downarrow$ & - & 0.015 & 0.021 & 0.019 & 0.019 & 0.008 & \textbf{0.006} & {\ul 0.007} & \textbf{0.006} \\
 & Macro F1 & - & 0.89 & 0.75 & 0.81 & 0.77 & 0.94 & 0.94 & {\ul 0.95} & \textbf{0.96} \\
 & Micro F1 & - & 0.90 & 0.85 & 0.87 & 0.86 & 0.94 & 0.94 & {\ul 0.95} & \textbf{0.96} \\
 & F1 & - & 0.91 & 0.83 & 0.87 & 0.86 & {\ul 0.96} & 0.93 & {\ul 0.96} & \textbf{0.97} \\
\hline
\end{tabular}%
}
\end{table*}

%% file: content/2-related-work.tex
A large number of methods have been proposed that train PGMs and NNs jointly. For example,  \citet{zhengConditionalRandomFields2015} proposed to combine conditional random fields (CRFs) and recurrent neural networks (RNNs), \citet{schwingFullyConnectedDeep2015, larssonMaxMarginLearningDeep2017, larssonProjectedGradientDescent2018, arnabHigherOrderConditional2016} showed how to combine CNNs and CRFs,  \citet{chenLearningDeepStructured2015} proposed to use densely connected graphical models with CNNs, and \citet{johnsonComposingGraphicalModels2017} combined latent graphical models with neural networks. 
\eat{Another virtue of DDNs is that they are easy to train and parallelizable, making them an attractive choice. As far as we know, ours is the first work that shows how to jointly train a dependency network, neural network hybrid.}
The combination of PGMs and NNs has also been applied to improve performance on a wide variety of real-world tasks. Notable examples include human pose estimation \cite{tompsonJointTrainingConvolutional2014, liangLimbBasedGraphicalModel2018, songThinSlicingNetworkDeep2017, yangEndtoEndLearningDeformable2016}, semantic labeling of body parts \cite{kirillovJointTrainingGeneric2016}, stereo estimation \cite{knobelreiterEndtoEndTrainingHybrid2017}, language understanding \cite{yaoRecurrentConditionalRandom2014}, \eat{joint intent detection and slot filling \cite{xuConvolutionalNeuralNetwork2013}, polyphonic piano music transcription \cite{sigtiaEndtoEndNeuralNetwork2016a},} face sketch synthesis \cite{zhuLearningDeepPatch2021}\eat{, sea ice floe segmentation \cite{nagiRUFEffectiveSea2021}} and crowd-sourcing aggregation \cite{liCrowdsourcingAggregationDeep2021}).
These hybrid models have also been used for solving a range of computer vision tasks such as semantic segmentation \cite{arnabConditionalRandomFields2018, guoSemanticImageSegmentation2021}, image crowd counting \cite{hanImageCrowdCounting2017}, \eat{Not a CV task structured output prediction \cite{pmlr-doNeuralConditionalRandom2010},} visual relationship detection \cite{yuProbabilisticGraphicalModel_2022_CVPR}, modeling for epileptic seizure detection \cite{craleyIntegratingConvolutionalNeural2019a}, \eat{Not a CV task phonetic recognition \cite{morrisConditionalRandomFields2008},} face sketch synthesis \cite{zhangNeuralProbabilisticGraphical2020}, semantic image segmentation \cite{chenDeepLabSemanticImage2018, linEfficientPiecewiseTraining2015}, 2D Hand-pose Estimation \cite{kongAdaptiveGraphicalModel2019}, depth estimation from a single monocular image \cite{liuDeepConvolutionalNeural2014}, animal pose tracking \cite{wuDeepGraphPoseNEURIPS2020} and pose estimation \cite{chenArticulatedPoseEstimation2014}. \eat{Not a CV task and adaptive team training \cite{rollArtificialIntelligenceEducation2021}} \eat{As far as we know, ours is the first work that uses jointly trained PGM+NN combinations to solve multi-label action (in videos) and image classification tasks.}

To date, dependency networks have been used to solve various tasks such as collective classification \cite{Neville2003CollectiveCW}, binary classification \cite{gamez2006dependency, gamez2008robust}, multi-label classification  \cite{guo2011multi}, part-of-speech tagging \cite{tarantola2002dependency}, relation prediction \cite{katzourisInductiveLogicProgramming2022b}, text classification \cite{guo_2020_deep} and collaborative filtering \cite{heckerman2000dependency}. However, DDNs have traditionally been restricted to Gibbs sampling \cite{heckerman2000dependency} and mean-field inference \cite{lowd_2011_mean}, showing limited compatibility with advanced probabilistic inference methods \cite{lowdClosedFormLearningMarkov2012a}. This study marks the inaugural attempt to incorporate advanced inference methods for the MPE task in DDNs, utilizing jointly trained networks for MLC scenarios.

%% file: content/7-conclusion.tex

\old{More and more state-of-the-art methods for challenging applications of computer vision tasks usually use deep neural networks. Deep neural networks are good at extracting features in vision tasks like image classification, video classification, object detection, image segmentation, and others. Nevertheless, for more complex tasks involving multi-label classification, these methods cannot model crucial information like inter-label dependencies. In this paper, we proposed a new modeling framework called deep dependency networks (DDNs) that combines a dependency network with a neural network and demonstrated via experiments, on three video and three image datasets, that it outperforms the baseline neural network, sometimes by a substantial margin. The key advantage of DDNs is that they explicitly model and reason about the relationship between the labels, and often improve model performance without considerable overhead. DDNs are also able to model additional relationships that are missed by other state-of-the-art methods that use transformers, attention module, and GAT. In particular, DDNs are simple to use, admit fast learning and inference, are easy to parallelize, and can leverage modern GPU architectures.}

\sva{More and more state-of-the-art methods for challenging applications of computer vision tasks usually use deep neural networks. Deep neural networks are good at extracting features in vision tasks like image classification, video classification, object detection, image segmentation, and others. Nevertheless, for more complex tasks involving multi-label classification, these methods cannot model crucial information like inter-label dependencies and infer about them. In this work, we present novel inference algorithms for Deep Dependency Networks (DDNs), a powerful neuro-symbolic approach, that exhibit consistent superiority compared to traditional neural network baselines, sometimes by a substantial margin. These algorithms offer an improvement over existing Gibbs Sampling-based inference schemes used for DDNs, without incurring significant computational burden. Importantly, they have the capacity to infer inter-label dependencies that are commonly overlooked by baseline techniques utilizing transformers, attention modules, and Graph Attention Networks (GAT). \eat{The algorithms are amenable to parallel processing and compatible with an array of optimization techniques, thereby facilitating integration with cutting-edge optimization software and faster inference on GPUs. } By formulating the inference procedure as an optimization problem, our approach permits the integration of domain-specific constraints, resulting in a more knowledgeable and focused inference process. In particular, our optimization-based approach furnishes a robust and computationally efficient mechanism for inference, well-suited for handling intricate multi-label classification tasks.}


Avenues for future work include: applying the setup described in the paper to other multi-label classification tasks in computer vision, natural language understanding, and speech recognition; converting DDNs to MRFs for better inference \cite{lowdClosedFormLearningMarkov2012a}; exploring and validating the neuro-symbolic benefits of DDNs such as improved accuracy in predictions and enhanced interpretability of model decisions; etc.


\eat{The framework provided here can be used as a feature processor for any multi-label classification task, providing potentially significant improvements over the baseline CNNs with small added overhead for training and inference. The method also does not depend on a particular feature extractor, allowing for the use of the current state-of-the-art methods with no modification. }


%% file: content/appendix.tex
\renewcommand\thesection{\Alph{section}}
\renewcommand\thesubsection{\thesection.\Alph{subsection}}
\setcounter{section}{0}

\renewcommand{\algorithmicrequire}{\textbf{Input:}}
\renewcommand{\algorithmicensure}{\textbf{Output:}}
\renewcommand{\thefigure}{SF\arabic{figure}} 
\renewcommand{\thetable}{ST\arabic{table}} 

\section{DERIVING THE OBJECTIVE FUNCTION FOR MOST PROBABLE EXPLANATIONS (MPE)}
\label{sec:supp_obj_as_opt}
Let us consider the scoring function we aim to maximize, given as follows:
\begin{equation}
\label{eq:sup_opt_1}
    \underset{\rvx}{\text{maximize}} \sum_{i=1}^{n} \log \left( P_i(x_{i} \mid \rvx_{-i},\rve)\right)
\end{equation}
Given that $x_i$ is binary, namely, $x_i \in \{0,1\}$, we can express the scoring function as follows:
\begin{equation}
\label{eq:sup_opt_2}
    \underset{\rvx}{\text{maximize}} \sum_{i=1}^{n} \Bigg( x_i \log \left( P_i(X_{i}=1 \mid \rvx_{-i},\rve) \right) + \, (1 - x_i) \log \left( 1 - P_i(X_{i}=1 \mid \rvx_{-i},\rve) \right) \Bigg)
\end{equation}

where
\[ P_i(X_i=1 | \rvx_{-i},\rve) = \sigma \left ( \sum_{j=1}^{|\rve|}w_{ij} e_j + \sum_{\substack{k=1 \\ k \neq i}}^{|\rvx|} v_{ik} x_k + b_i\right )\]
Let $p_i = P_i(X_i=1 | \rvx_{-i},\rve)$ and $z_i=\sum_{j=1}^{|\rve|}w_{ij} e_j + \sum_{\substack{k=1 \\ k \neq i}}^{|\rvx|} v_{ik} x_k + b_i$. Then, the MPE task, which involves optimizing the scoring function given above, can be expressed as:
\begin{align}
& \underset{\rvx, \rvz}{\text{maximize}} \sum_{i=1}^n \left( x_i \log p_i + (1 - x_i) \log(1 - p_i) \right) \label{eq:sup_opt_ilp_objective} \\
& \text{subject to:} \notag \\
& p_i = \sigma(z_i), \quad \forall i \in \{1,\ldots,n\} \label{eq:sup_opt_ilp_constraint1} \\
& z_i = \sum_{j=1}^{|\rve|}w_{ij} e_j + \sum_{\substack{k=1 \\ k \neq i}}^{|\rvx|} v_{ik} x_k + b_i, \quad \forall i \in \{1,\ldots,n\} \label{eq:sup_opt_ilp_constraint2} \\
& x_i \in \{0, 1\}, \quad \forall i \in \{1,\ldots,n\} \label{eq:sup_opt_ilp_constraint3}
\end{align}

The first constraint in the formulation, as expressed by \ref{eq:sup_opt_ilp_constraint1}, pertains to the sigmoid function employed in the logistic regression module. The subsequent constraint, defined by \ref{eq:sup_opt_ilp_constraint2}, formalizes the product between the weights and inputs in the Dependency Network (DN). Here, $ e_i $ represents evidence values supplied to the DN, while $ \rvx $ represents the decision variables that are subject to optimization. Finally, \ref{eq:sup_opt_ilp_constraint3} specifies that the input variables must be constrained to integer values of 0 or 1.

The assembled constraints collectively simulate a forward pass through the network. The objective function is designed to maximize the scoring function pertinent to the Most Probable Explanation (MPE). This framework adeptly aligns the optimization task to maximize the MPE score.

The substitution of the constraint $ P_i(x_i | \rvx_{-i},\rve) = \sigma(z_i) = \frac{1}{1 + e^{(- z_{i})}} $ into the objective function and the subsequent algebraic simplification result in a simplified formulation as follows:

\begin{equation}
\label{eq:sup_simplify_opt_function}
\begin{aligned}
& x_i \log p_i + (1 - x_i) \log (1 - p_i) \\
& =x_i \log \left( \frac{e^{z_i}}{1 + e^{z_i}} \right) + (1 - x_i) \log \left(1 - \frac{e^{z_i}}{1 + e^{z_i}} \right) \\
& =x_i \log \left( \frac{e^{z_i}}{1 + e^{z_i}} \right) + (1 - x_i) \log \left( \frac{1}{1 + e^{z_i}} \right)  \\
& =x_i \log e^{z_i} - x_i \log (1 + e^{z_i}) - \log (1 + e^{z_i}) + x_i \log (1 + e^{z_i}) \\
& =x_i \log e^{z_i} - \log (1 + e^{z_i})  \\
& =x_i z_i - \log (1 + e^{z_i}) 
\end{aligned}
\end{equation}

Substituting \eqref{eq:sup_simplify_opt_function} in the objective function of our optimization problem, we get
\begin{align}
&\underset{\rvx, \rvz}{\text{maximize}} \sum_{i=1}^n x_i z_{i} - \log \left(1 + e^{z_{i}}\right) \label{eq:sup_opt_ilp_obj} \\
&\text{subject to:} \nonumber \\
&z_i = \sum_{j=1}^{|\rve|}w_{ij} e_j + \sum_{\substack{k=1 \\ k \neq i}}^{|\rvx|} v_{ik} x_k + b_i, \quad \forall i \in \{1,\ldots,n\} \label{eq:sup_opt_ilp_z} \\
&x_i \in \{0, 1\}, \quad \forall i \in \{1,\ldots,n\} \label{eq:sup_opt_ilp_x}
\end{align}

\subsection{Utilizing Piecewise Linear Approximation for Non-linear Functions}

\begin{figure}[ht!]
\centering
\caption{Piece-wise linear approximation of $log(1+e^{z_i})$}
\includegraphics[width=0.7\textwidth]{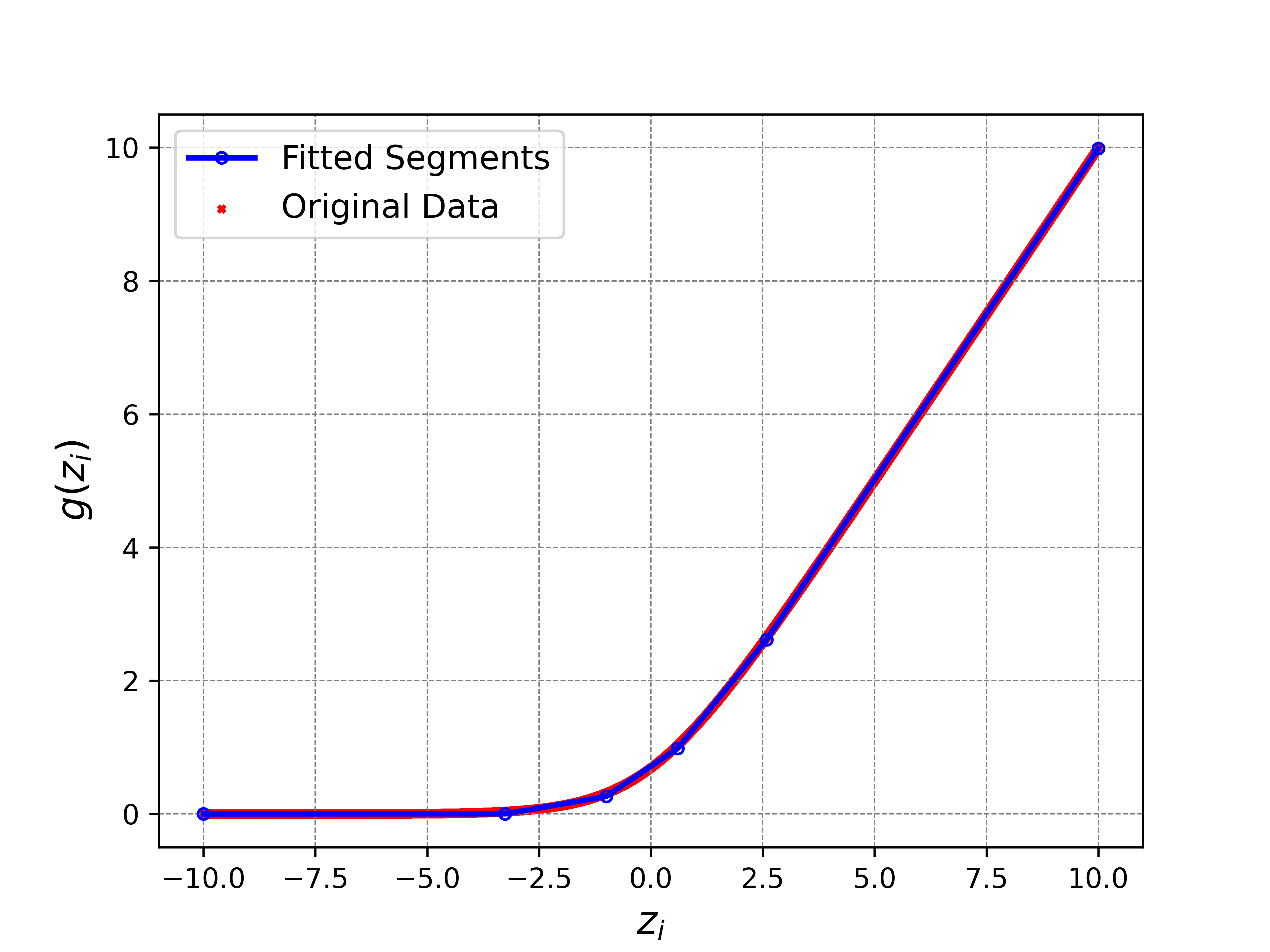}
\label{fig:supp_piecewise_linear}
\end{figure}

The objective function expressed in equation \eqref{eq:sup_opt_ilp_obj} is non-linear due to the inclusion of the term $ \log(1+e^{z_i}) $. $f(z)$ can be used as a piecewise approximation  for $\log(1+e^{z_i})$. 

\begin{equation}
    f(z) = \begin{cases} z & z \gg 1 \\ e^{z} & z \ll 1 \\\log(1 + e^z) & \text{otherwise} \end{cases}
\end{equation}

To obtain a piecewise linear approximation of $ \log(1+e^{z_i}) $, a single linear function suffices for $ z \gg 1 $, while the majority of the linear pieces are utilized for approximating the function near 1. A piecewise linear approximation of the function is detailed in the figure \ref{fig:supp_piecewise_linear}. We employ five segments for the approximation. These piecewise functions can be integrated as linear constraints, facilitating the conversion of the non-linear objective into a linear objective with ease. We also present the piecewise equations for the approximation as follows - 
\begin{table}[ht]
\centering
\caption{Piecewise Linear Approximation for \\ $ g(z) \approx \log(1+e^{z_i}) $}
\label{tab:equations}
\begin{tabular}{ll}
\toprule
$z$ & $g(z)$ \\
\midrule
$]-\infty, -3.257)$ & $0$ \\
$[-3.257, -0.998)$ & $(2^{-4} + 2^{-5} + 2^{-6} + 2^{-8})z + 0.379$ \\
$[-0.998, 0.602)$ & $(2^{-2} + 2^{-3} + 2^{-4} + 2^{-7} + 2^{-8})z + 0.715$ \\
$[0.602, 2.584)$ & $(2^{-1} + 2^{-2} + 2^{-4} + 2^{-7})z + 0.492$ \\
$[2.584, +\infty[$ & $z$ \\
\bottomrule
\end{tabular}
\end{table}

Thus after replacing $\log(1+e^{z_i})$ with its piece-wise approximation ($g(z)$) we get
\begin{equation}
\label{eq:supp_opt_final}
\begin{aligned}
&\underset{\rvx, \rvz, \alpha}{\text{maximize}} \sum_{i=1}^n x_i z_i - g(z_i) \\
&\text{subject to:} \\
& z_i = \sum_{j=1}^{|\rve|} w_{ij} e_j + \sum_{\substack{k=1 \\ k \neq i}}^{|\rvx|} v_{ik} x_k + b_i, \quad \forall i \in \{1,\ldots,n\} \\
& x_i \in \{0, 1\}, \quad \forall i \in \{1,\ldots,n\} \\
& g(z_i) = \alpha_{1i} \times 0 + \alpha_{2i} \times ((2^{-4} + 2^{-5} + 2^{-6} + 2^{-8})z_i + 0.379) + \\
& \qquad\qquad\qquad \alpha_{3i} \times ((2^{-2} + 2^{-3} + 2^{-4} + 2^{-7} + 2^{-8})z_i + 0.715) + \\
& \qquad\qquad\qquad \alpha_{4i} \times ((2^{-1} + 2^{-2} + 2^{-4} + 2^{-7})z_i + 0.492) + \\
& \qquad\qquad\qquad \alpha_{5i} z_i, \quad \forall i \in \{1,\ldots,n\} \\
& \sum_{j=1}^{5} \alpha_{ji} = 1, \quad \forall i \in \{1,\ldots,n\} \\
& \alpha_{ji} \in \{0, 1\}, \quad \forall j \in \{1,2,3,4,5\}, \forall i \in \{1,\ldots,n\} \\
& \alpha_{1i} = 1 \Rightarrow z_i < -3.257, \quad \forall i \in \{1,\ldots,n\} \\
& \alpha_{2i} = 1 \Rightarrow -3.257 \leq z_i < -0.998, \quad \forall i \in \{1,\ldots,n\} \\
& \alpha_{3i} = 1 \Rightarrow -0.998 \leq z_i < 0.602, \quad \forall i \in \{1,\ldots,n\} \\
& \alpha_{4i} = 1 \Rightarrow 0.602 \leq z_i < 2.584, \quad \forall i \in \{1,\ldots,n\} \\
& \alpha_{5i} = 1 \Rightarrow z_i \geq 2.584, \quad \forall i \in \{1,\ldots,n\}
\end{aligned}
\end{equation}

The utilization of binary variables, $ \alpha_{ji}$, allows for a piece-wise linear approximation of the logarithm of $ (1+e^{z_i}) $, with these variables serving as selectors to determine the appropriate linear segment based on the value of $ z_i $. It is worth noting that the final five constraints are indicator constraints, which can be linearized using the big-M method as described in \citet{griva_2009_linear}. Consequently, the problem can be formulated as a mixed-integer multi-linear optimization problem. The mixed-integer multi-linear problem can be further converted to a mixed-integer linear problem (MILP) using the approach describe next.

\section{FORMULATING LOGICAL AND BETWEEN BINARY VARIABLES AS LINEAR CONSTRAINTS}
Consider an optimization problem involving three binary variables $X_1$, $X_2$, and $X_3$, where $x_1$, $x_2$, and $x_3$ denote their respective assignments. We define an objective function that incorporates products of these binary variables as follows - 
\begin{equation}
    \max_{x_1, x_2, x_3} \; x_1x_2 - x_2x_3 + x_1x_3
\end{equation}
Although constraints are not incorporated in this instance, the same methodology can be extended to constrained optimization problems. Subsequently, auxiliary variables $ z_1 $, $ z_2 $, and $ z_3 $ are introduced to account for each of the binary products. The optimization problem can be stated as follows:

\begin{equation}
\begin{aligned}
    \max_{x_1, x_2, x_3} \quad & z_1 - z_2 + z_3 \\
    \text{s.t.} \quad & x_1 \land x_2 = z_1, \\
              & x_2 \land x_3 = z_2, \\
              & x_1 \land x_3 = z_3.
\end{aligned} 
\end{equation}

The optimization problem may be expressed by incorporating additional linear constraints to represent each boolean product. The resulting problem is presented as follows:

\begin{equation}
\begin{aligned}
    \max_{x_1, x_2, x_3} \quad & z_1 - z_2 + z_3 \\
    \text{s.t.} \quad & x_1 + x_2 - 1 \leq z_1, \\
              & z_1 \leq x_1, \\
              & z_1 \leq x_2, \\
              & x_2 + x_3 - 1 \leq z_2, \\
              & z_2 \leq x_2, \\
              & z_2 \leq x_3, \\
              & x_1 + x_3 - 1 \leq z_3, \\
              & z_3 \leq x_1, \\
              & z_3 \leq x_3.
\end{aligned}
\end{equation}

The problem formulation outlined in Section \ref{sec:supp_obj_as_opt}, and more specifically, equation \eqref{eq:supp_opt_final}, includes expressions of the form $ x_i x_j $. These terms serve to represent the logical AND operation between binary variables $ x_i $ and $ x_j $, and are a direct result of the product $ x_i \times g_i $. Consequently, this optimization problem is classified as a mixed-integer multi-linear optimization problem.

The formulation detailed in this section enables the capture of these logical AND operations between binary variables through linear constraints, thereby enabling the utilization of Mixed-Integer Linear Programming (MILP) solvers, e.g., \citet{gurobi}, to find the optimal solution (or an anytime, near optimal solution if a time bound is specified).


\section{GIBBS SAMPLING FOR DDNS}

\begin{algorithm}[ht!]
    \caption{Gibbs Sampling for DDNs}
    \label{alg:gibbs}
    \begin{algorithmic}[1]
        \Require video segment/image $\rvv$, number of samples $N$, DDN $\langle \mathcal{N},\mathcal{D} \rangle$
        \Ensure An estimate of the marginal probability distribution over each label $X_i$ of the DDN given $\rvv$
        \State $\rve \gets \mathbb{N}(\rvv)$
        \State Randomly initialize $\rmX = \rvx^{(0)}$
        \For{$j = 1$ \textbf{to} $N$}
            \State $\pi \gets$ Generate random permutation of $[1,n]$
            \For{$i = 1$ \textbf{to} $n$}
                \State $x_{\pi(i)}^{(j)} \sim P_{\pi(i)}(x_{\pi(i)} | \rvx_{\pi(1):\pi(i-1)}^{(j)}, \rvx_{\pi(i+1):\pi(n)}^{(j-1)}, \rve)$
            \EndFor
        \EndFor
        \For{$i = 1$ \textbf{to} $n$}
            \State $\hat{P}_i(x_i|\rvv) = \frac{1}{N} \sum_{j=1}^{N} P_i(x_i|\rvx_{-i}^{(j)},\rve)$
        \EndFor
        \State \textbf{return} $\{\hat{P}_i(x_i|\rvv)|i \in \{1,\ldots,n\}\}$
    \end{algorithmic}
\end{algorithm}

This section describes the Gibbs Sampling inference procedure for DDNs (see Algorithm \ref{alg:gibbs}). Gibbs Sampling serves as an approximation method for the Most Probable Explanation (MPE) inference task in DDNs by offering max-marginals, which can subsequently be utilized to approximate MPE. The inputs to the algorithm are (1) a video segment/image $\rvv$, (2) the number of samples $N$ and (3) trained DDN model $\langle \mathcal{N}, \mathcal{D}\rangle$. The algorithm begins (see step 1) by extracting features $\rve$ from the video segment/image $\rvv$ by sending the latter through the neural network $\mathcal{N}$ (which represents the function $\mathbb{N}$). Then in steps 2--8, it generates $N$ samples via Gibbs sampling. The Gibbs sampling procedure begins with a random assignment to all the labels (step 2). Then at each iteration (steps 3--8), it first generates a random permutation $\pi$ over the $n$ labels and samples the labels one by one along the order $\pi$ (steps 5--7). To sample a label indexed by $\pi(i)$ at iteration $j$, we  compute $P_{\pi(i)}(x_{\pi(i)} | \rvx_{\pi(1):\pi(i-1)}^{(j)}, \rvx_{\pi(i+1):\pi(n)}^{(j-1)}, \rve)$ from the DN $\mathcal{D}$ where $\rvx_{\pi(1):\pi(i-1)}^{(j)}$ and $\rvx_{\pi(i+1):\pi(n)}^{(j-1)}$ denote the assignments to all labels ordered before $x_{\pi(i)}$ at iteration $j$ and the assignments to all labels ordered after $x_{\pi(i)}$ at iteration $j-1$ respectively.

After $N$ samples are generated via Gibbs sampling, the algorithm uses them to estimate (see steps 9--11) the (posterior) marginal probability distribution at each label $X_i$ given $\rvv$ using the mixture estimator \cite{liubook}. The algorithm terminates (see step 12) by returning these posterior estimates.

\eat{the end of this procedure we have N samples and now we need to estimate the marginal probability
distribution of each label $X_i$ which we do by performing the mixture estimator given in \eqref{eq:Mixture_Estimator1}. Finally the marginal probability distribution is returned for all the labels.}

\section{PRECISION BASED EVALUATION METRICS}

In Tables \ref{tab:supp-precision-mlac} and \ref{tab:supp-precision-mlic}, we present a comprehensive account of precision-focused metrics, specifically Mean Average Precision (mAP) and Label Ranking Average Precision (LRAP). It should be noted that these metrics are not directly applicable to the Most Probable Explanation (MPE) task, as they necessitate scores (probabilities) of the output labels. However, given that SlowFast, InceptionV3, Q2L, MSRN, and DDN-GS can provide such probabilities, their precision scores are anticipated to be elevated. We employed Gibbs Sampling to approximate the MPE, enabling the derivation of marginal probabilities to approximate max-marginals and, thereby, the MPE.

In majority of the instances, the Gibbs Sampling-based inference on DDNs yields performance metrics closely aligned with baseline methods. Variability exists, with some metrics exceeding the baseline in the context of MLAC and falling short in the case of MLIC. Notably, the Integer Linear Programming (ILP)-based approach for DDN outperforms all alternative methods across three datasets when evaluated on LRAP. DRF-based methods and DDN-specific inference techniques generally underperform on these precision metrics. Again, this outcome is anticipated, given that apart from Gibbs Sampling, none of the other techniques can generate probabilistic scores for labels, leading to their inferior performance.

\input{content/supp_mlac_metrics}
\input{content/supp_mlic_metrics}

\section{ANNOTATIONS COMPARISON BETWEEN Q2L AND DDN-MLP-JOINT ON THE MS-COCO DATASET}
\icml{
Table \ref{tab:annotation_comparison_supp} presents a qualitative evaluation of the label predictions produced by our DDN-ILP inference scheme compared to the baseline Q2L using randomly selected images from the MS-COCO dataset. This analysis provides a perspective on the advantages of DDN-ILP over Q2L, particularly in terms of correcting errors generated by the feature extractor. In the initial set of seven rows, DDN-ILP adds additional correct labels to the Q2L output. For certain images, the objects overlooked by Q2L are inherently challenging to identify. In these instances, the DDN-ILP method, which infers labels based on label relationships, effectively rectifies the limitations of Q2L. DDN-ILP refines Q2L's predictions in the following seven instances by removing incorrect predictions and aligning precisely with the ground truth labels. Finally, we examine cases where DDN-ILP's modifications result in inaccuracies (rows have grey background), contrasting with Q2L's correct predictions. These results demonstrate the effectiveness of DDN-ILP in improving the predicted labels relative to the baseline.}

\input{content/supp_qualitative}

%% file: content/supp_mlac_metrics.tex
\begin{table}[!ht]
\centering
\caption{Comparison of our methods with the feature extractor for MLAC - Precision-based metrics. The best/second best values are bold/underlined.}
\label{tab:supp-precision-mlac}
\begin{tabular}{@{}cllllll@{}}
\toprule
Method                                       & \multicolumn{2}{c}{Charades}  & \multicolumn{2}{c}{TACoS}     & \multicolumn{2}{c}{Wetlab}    \\ \cmidrule(l){2-7} 
                                             & mAP           & LRAP          & mAP           & LRAP          & mAP           & LRAP          \\ \midrule
SlowFast \cite{fan2020pyslowfast}           & \uline{0.39}  & \uline{0.53}  &               &               &               &               \\
InceptionV3 \cite{szegedyRethinkingInceptionArchitecture2016} &               &               & \uline{0.70}  & 0.81          & \uline{0.79}  & 0.82          \\
DRF - GS                                     & 0.27          & 0.44          & 0.56          & 0.79          & 0.54          & 0.76          \\
DRF - ILP                                    & 0.19          & 0.28          & 0.40          & 0.67          & 0.63          & 0.73          \\
DRF - IJGP                                   & 0.31          & 0.44          & 0.56          & 0.81          & 0.79          & \uline{0.85} \\
DDN - GS                                     & \textbf{0.40} & \textbf{0.55} & \textbf{0.75} & \uline{0.84}  & \textbf{0.84} & \textbf{0.87} \\
DDN - RW                                     & 0.19          & 0.28          & 0.49          & 0.66          & 0.63          & 0.77          \\
DDN - Greedy                                 & 0.32          & 0.30          & 0.50          & 0.69          & 0.65          & 0.78          \\
DDN - ILP                                    & 0.36          & \uline{0.53}  & 0.51          & \textbf{0.85} & 0.65          & \textbf{0.87} \\ \bottomrule
\end{tabular}
\end{table}

%% file: content/supp_mlic_metrics.tex
\begin{table}[ht]
\centering
\caption{Comparison of our methods with the feature extractor for MLIC for Precision based metrics. The best/second best values are bold/underlined.}
\label{tab:supp-precision-mlic}
\begin{tabular}{@{}ccccccc@{}}
\toprule
Method                                                    & \multicolumn{2}{c}{MS-COCO}   & \multicolumn{2}{c}{NUS-WIDE}  & \multicolumn{2}{c}{PASCAL-VOC} \\ \cmidrule(l){2-7} 
                                                          & mAP           & LRAP          & mAP           & LRAP          & mAP            & LRAP          \\ \midrule
Q2L \cite{liuQuery2LabelSimpleTransformer2021}          & \textbf{0.91} & \textbf{0.96} &               &               &                &               \\
MSRN \cite{quMultilayeredSemanticRepresentation2021}     &               &               & \textbf{0.62} & \textbf{0.85} & \uline{0.96}   & \textbf{0.98} \\
DRF - GS                                                  & 0.75          & 0.86          & 0.40          & 0.74          & 0.77           & 0.93          \\
DRF - ILP                                                 & 0.74          & 0.82          & 0.25          & 0.59          & 0.81           & 0.88          \\
DRF - IJGP                                                & 0.74          & 0.90          & 0.41          & 0.75          & 0.83           & 0.94          \\
DDN - GS                                                  & 0.84          & \uline{0.93}  & \uline{0.50}  & \uline{0.82}  & 0.92           & \uline{0.96}  \\
DDN - RW                                                  & 0.74          & 0.82          & 0.24          & 0.61          & 0.91           & 0.94          \\
DDN - Greedy                                              & 0.74          & 0.83          & 0.35          & 0.68          & 0.93           & 0.95          \\
DDN - ILP                                                 & \uline{0.85}  & 0.83          & 0.48          & \uline{0.82}  & \textbf{0.97}  & \textbf{0.98} \\ \bottomrule
\end{tabular}
\end{table}

%% file: content/supp_qualitative.tex
\begin{longtable}{|p{0.38\textwidth}|p{0.16\textwidth}|p{0.18\textwidth}|p{0.18\textwidth}|}
\caption{Comparison of labels predicted by Q2L~\cite{liuQuery2LabelSimpleTransformer2021} and our DDN-ILP scheme on the MS-COCO dataset. Labels inside \textbf{[]} represent the difference between the predictions of the two methods, assuming that a threshold of 0.5 is used (i.e., every label whose probability $ > 0.5$ is considered a predicted label). Due to the MPE focus in DDN-ILP, only label configurations are generated, omitting corresponding probabilities. }
\label{tab:annotation_comparison_supp}\\
\hline
\textbf{Image} & \textbf{Ground Truth} & \textbf{Q2L} & \textbf{DDN} \\ \hline
\endfirsthead
\endhead
%
\parbox[c]{1em}{\includegraphics[width=0.38\textwidth]{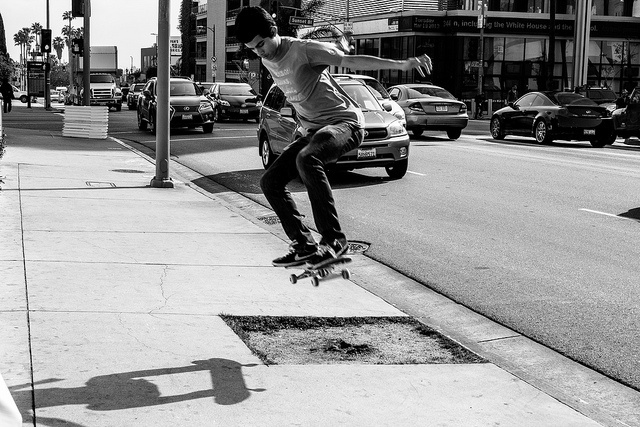}} & person, car, truck, traffic light, skateboard                  & person (1.00), car (1.00), skateboard (1.00), \textbf{[truck (0.33), traffic light (0.41)]}                                   & person, car, truck, traffic light, skateboard        \\ \hline
\parbox[c]{1em}{\includegraphics[width=0.38\textwidth]{images/ddn_vs_cnn/ddn_better/COCO_val2014_000000048905.jpg}} & knife, spoon, microwave, oven, sink, refrigerator              & microwave (1.00), oven (1.00), sink (1.00), refrigerator (1.00), knife (0.98), \textbf{[spoon (0.38)]}                        & knife, spoon, microwave, oven, sink, refrigerator    \\ \hline
\parbox[c]{1em}{\includegraphics[width=0.38\textwidth]{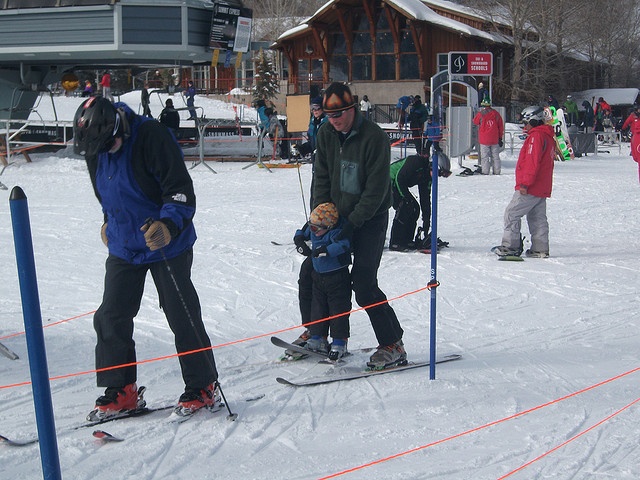}} & person, skis, snowboard                                        & person (1.00), skis (1.00), \textbf{[snowboard (0.20)]}                                                                       & person, skis, snowboard                              \\ \hline
\parbox[c]{1em}{\includegraphics[width=0.38\textwidth]{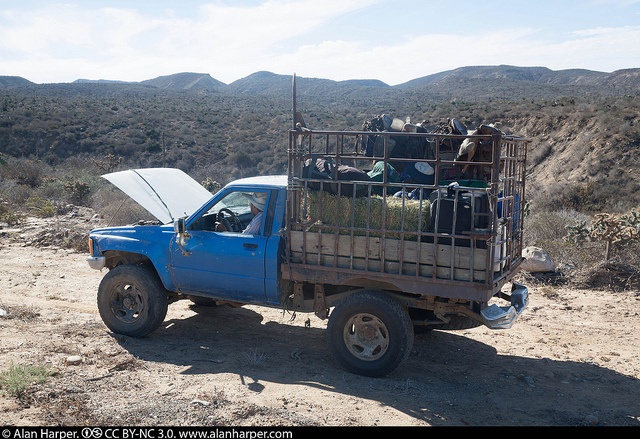}} & person, truck, suitcase                                        & truck (1.00), suitcase (0.98), \textbf{[person (0.24)]}                                                                       & person, truck, suitcase                              \\ \hline
\parbox[c]{1em}{\includegraphics[width=0.38\textwidth]{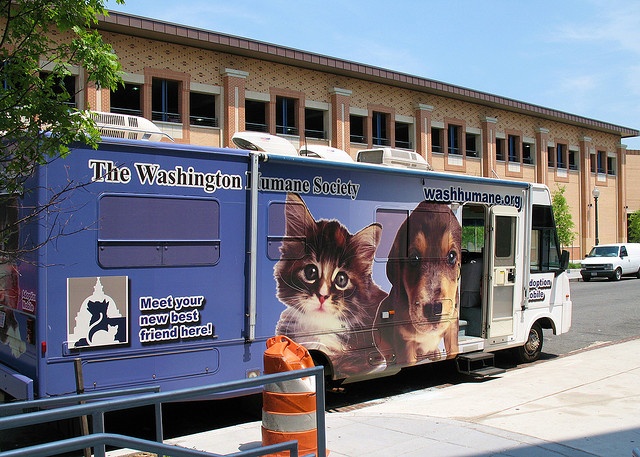}} & car, bus, truck, cat, dog                                      & car (0.99), truck (0.99), cat (0.96), \textbf{[bus (0.37), dog (0.15)]}                                                       & car, bus, truck, cat, dog                            \\ \hline
\parbox[c]{1em}{\includegraphics[width=0.38\textwidth]{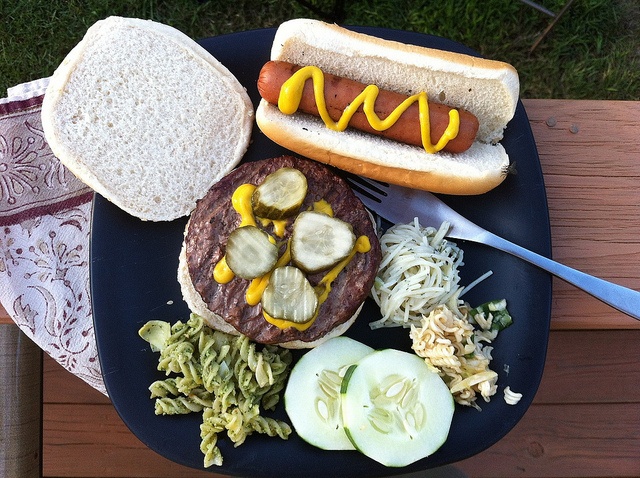}} & fork, sandwich, hot dog, dining table                          & fork (1.00), hot dog (1.00), dining table (0.90), \textbf{[sandwich (0.28)]}                                                  & fork, sandwich, hot dog, dining table                \\ \hline
\parbox[c]{1em}{\includegraphics[width=0.38\textwidth]{images/ddn_vs_cnn/ddn_better/COCO_val2014_000000495125.jpg}} & person, bottle, wine glass, chair, dining table                & person (1.00), bottle (1.00), wine glass (1.00), dining table (0.99), \textbf{[chair (0.25)] }                                & person, bottle, wine glass, chair, dining table      \\ \hline
\parbox[c]{1em}{\includegraphics[width=0.38\textwidth]{images/ddn_vs_cnn/ddn_better/COCO_val2014_000000516038.jpg}} & person, sports ball, baseball glove,                           & person (1.00), baseball glove (1.00), \textbf{[sports ball (0.27)]}                                                           & person, sports ball, baseball glove,                 \\ \hline
\parbox[c]{1em}{\includegraphics[width=0.38\textwidth]{images/ddn_vs_cnn/ddn_better/COCO_val2014_000000145020.jpg}} & person, teddy bear                                             & person (1.00), teddy bear (1.00), \textbf{[handbag (0.58), clock (0.96)]}                                                     & person, teddy bear                                   \\ \hline
\parbox[c]{1em}{\includegraphics[width=0.38\textwidth]{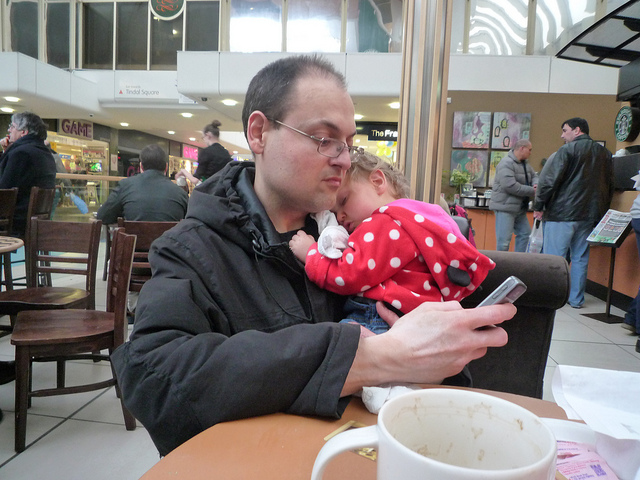}} & person, cup, chair, dining table, cell phone,                  & person (1.00), cup (1.00), chair (1.00), dining table (1.00), cell phone (1.00), \textbf{[clock (0.92), potted plant (0.85)]} & person, cup, chair, dining table, cell phone,        \\ \hline
\parbox[c]{1em}{\includegraphics[width=0.38\textwidth]{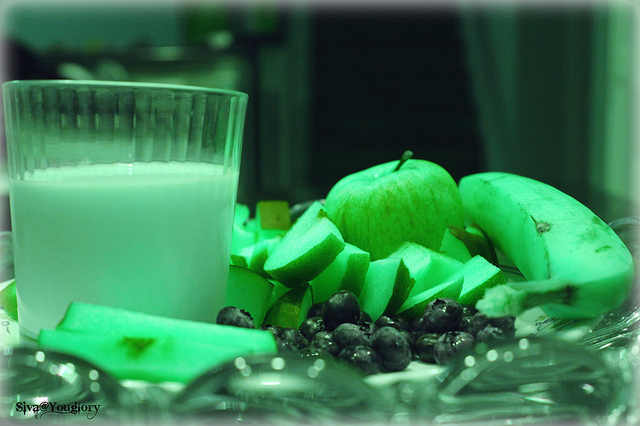}} & cup, banana, apple                                             & cup (1.00), banana (1.00), apple (1.00), \textbf{[spoon (0.90), fork (0.55)]}                                                 & cup, banana, apple                                   \\ \hline
\parbox[c]{1em}{\includegraphics[width=0.38\textwidth]{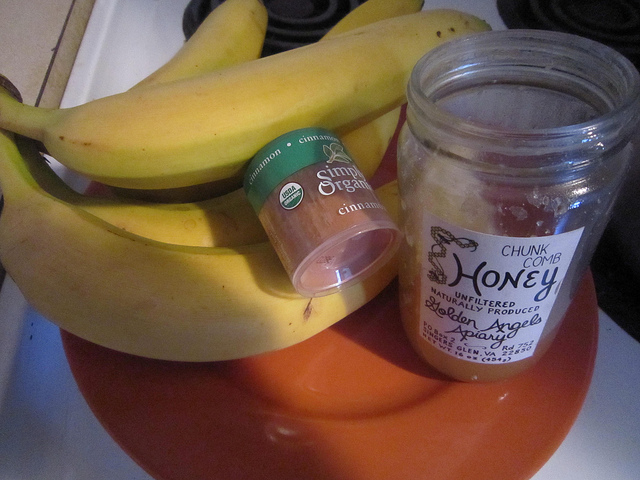}} & cup, banana                                                    & banana (1.00), \textbf{[bottle (0.90), oven (0.66)]},  cup (0.54)                                                             & cup, banana                                          \\ \hline
\parbox[c]{1em}{\includegraphics[width=0.38\textwidth]{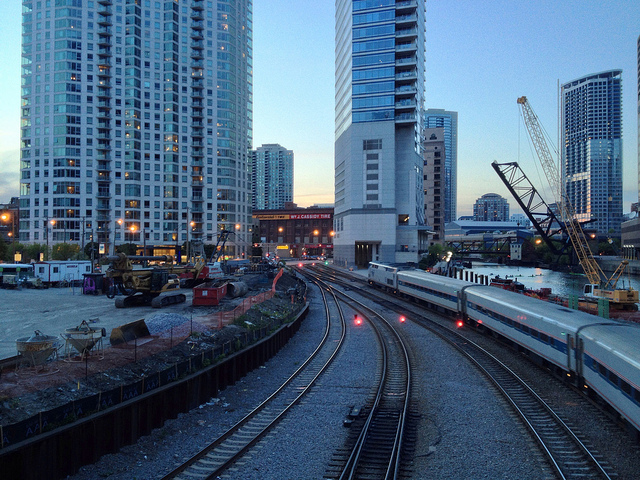}} & train                                                          & train (1.00), \textbf{[traffic light (0.90), car (0.56), , truck (0.51)]}                                                     & train                                                \\ \hline
\parbox[c]{1em}{\includegraphics[width=0.38\textwidth]{images/ddn_vs_cnn/ddn_better/COCO_val2014_000000150843.jpg}} & person, bowl, oven                                             & person (1.00),  oven (1.00) , bowl (0.99),  \textbf{[chair (0.85), spoon (0.59)]}                                             & person, bowl, oven                                   \\ \hline
\parbox[c]{1em}{\includegraphics[width=0.38\textwidth]{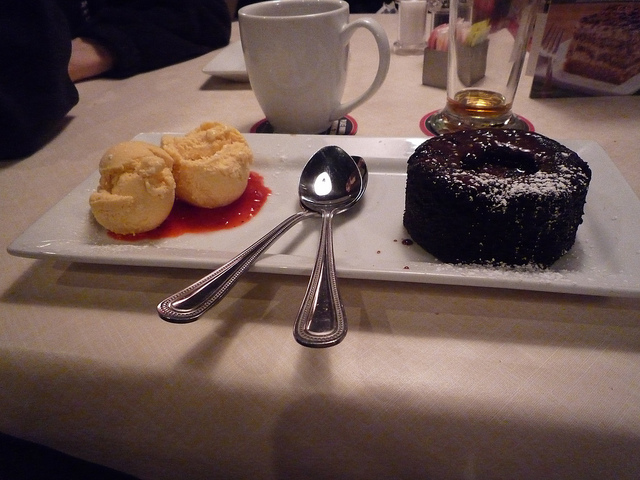}} & person, cup, spoon, cake, dining table                         & cup (1.00), spoon (1.00), cake (1.00), dining table (0.95), person (0.91), \textbf{[donut (0.89), fork (0.65)]}               & person, cup, spoon, cake, dining table               \\ \hline
\rowcolor[HTML]{EFEFEF} 
\parbox[c]{1em}{\includegraphics[width=0.38\textwidth]{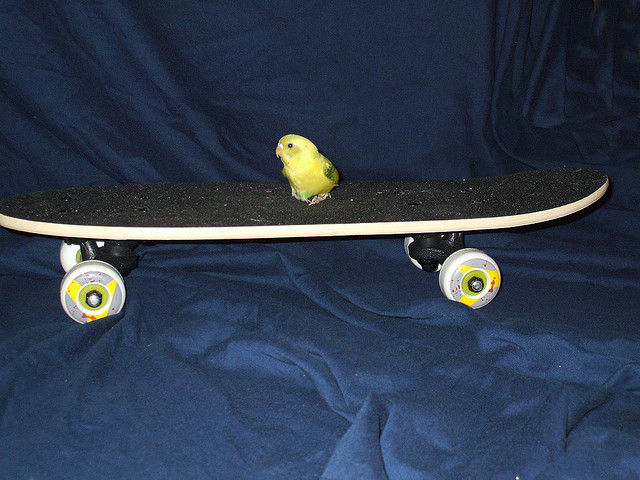}}  & bird, skateboard, couch                                        & bird (1.00), skateboard (1.00), couch (0.73)                                                                         & bird, skateboard, \textbf{[couch]}                                   \\ \hline
\rowcolor[HTML]{EFEFEF} 
\parbox[c]{1em}{\includegraphics[width=0.38\textwidth]{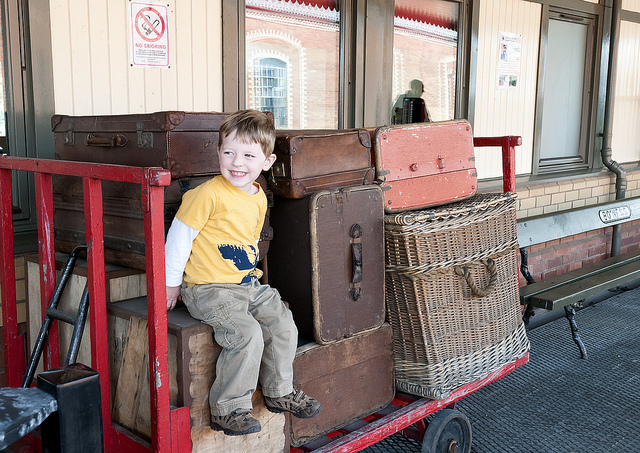}}  & person, bench, suitcase                                          & person (1.00), suitcase (1.00), bench (0.57)                                                                         & person, suitcase, \textbf{[bench]}                                          \\ \hline
\rowcolor[HTML]{EFEFEF} 
\parbox[c]{1em}{\includegraphics[width=0.38\textwidth]{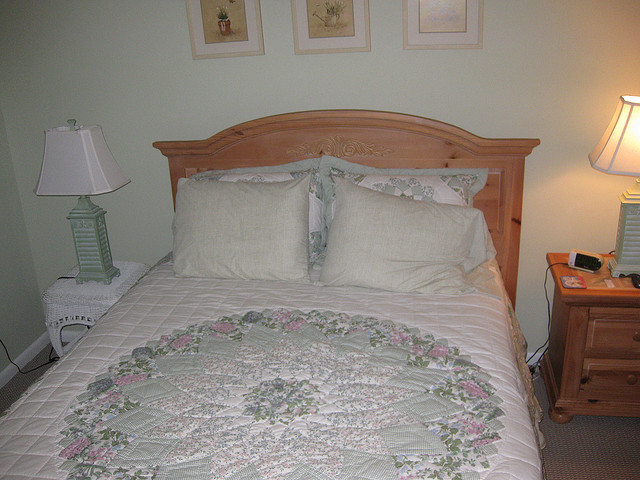}}  & bed, clock                                                       & bed (1.00), clock (0.51)                                                                                             & bed, \textbf{[clock]}                                                      \\ \hline
\rowcolor[HTML]{EFEFEF} 
\parbox[c]{1em}{\includegraphics[width=0.38\textwidth]{images/ddn_vs_cnn/ddn_worse/COCO_val2014_000000010363.jpg}}  & bicycle, car, cat, bottle                                        & bicycle (1.00), car (1.00), cat (1.00), bottle   (0.77)                                                              & bicycle, car, cat, \textbf{[bottle]}                                      \\ \hline
\rowcolor[HTML]{EFEFEF} 
\parbox[c]{1em}{\includegraphics[width=0.38\textwidth]{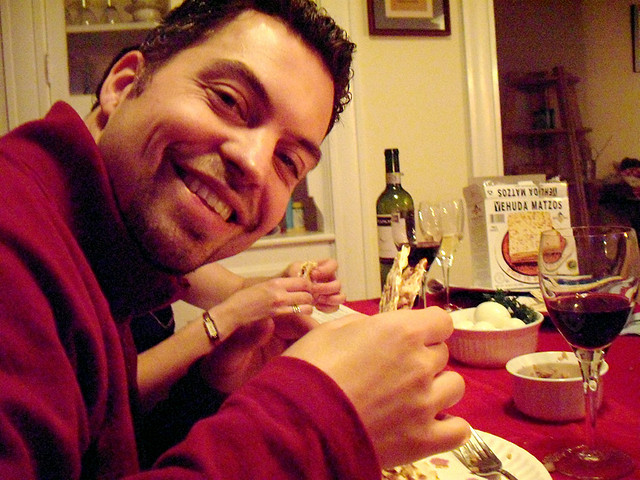}}  & person, bottle, wine glass, fork, bowl, dining   table, sandwich & person (1.00), bottle (1.00), wine glass   (1.00), fork (1.00), bowl (1.00), dining table (0.99), sandwich (0.52)    & person, bottle, wine glass, fork, bowl, dining   table, \textbf{[sandwich]} \\ \hline

\end{longtable}